\def\uselighterheadingblue{}
\def\abstractboxbackground{abstractappendixblue}
\def\abstractboxframe{abstractappendixblue}
\def\abstractseparator{}
\documentclass[]{inkmind}

\usepackage{titletoc}       
\usepackage{url}            
\usepackage{amsfonts}       
\usepackage{nicefrac}       
\usepackage{colortbl}       
\usepackage{amsmath}
\usepackage{wrapfig}
\usepackage{algorithm}
\usepackage{algpseudocode}
\usepackage{listings}
\definecolor{promptbg}{HTML}{FAFAFA}
\definecolor{promptframe}{HTML}{454545}
\definecolor{prompttitle}{HTML}{3F3F3F}
\lstdefinestyle{promptstyle}{
  basicstyle=\ttfamily\fontsize{6.6}{7.3}\selectfont,
  breaklines=true,
  breakatwhitespace=false,
  columns=fullflexible,
  keepspaces=true,
  showstringspaces=false,
  showspaces=false,
  showtabs=false,
  tabsize=2,
  frame=none,
  xleftmargin=0pt,
  xrightmargin=0pt,
  aboveskip=0pt,
  belowskip=0pt
}
\newtcblisting{promptbox}[1]{
  enhanced,
  breakable,
  listing only,
  listing options={style=promptstyle},
  title={#1},
  colback=promptbg,
  colframe=promptframe,
  colbacktitle=prompttitle,
  coltitle=white,
  fonttitle=\rmfamily\normalsize,
  boxrule=0.65pt,
  arc=2mm,
  outer arc=2mm,
  left=4mm,
  right=4mm,
  top=2mm,
  bottom=2mm,
  lefttitle=4mm,
  righttitle=4mm,
  toptitle=0.9mm,
  bottomtitle=0.9mm,
  before skip=0.5em,
  after skip=0.7em,
  segmentation style={solid,draw=promptframe,line width=0.5pt}
}

\title{VISD: Enhancing Video Reasoning via Structured Self-Distillation}

\author[1,*]{Hao Lin}
\author[2,*]{Kunyang Lv}
\author[3,5]{Xu Jiang}
\author[4,5]{Jingqi Tian}
\author[3,5]{Zhongjing Du}
\author[3,5]{Jiayu Ding}
\author[3,5]{Qiaoman Zhang}
\author[3,5,\dagger]{Hongbo Jin}

\affiliation[1]{HUST}
\affiliation[2]{Wuhan University}
\affiliation[3]{Peking University}
\affiliation[4]{Tsinghua University}
\affiliation[5]{InkMind.AI}

\contribution[*]{Equal contribution,}
\contribution[\dagger]{Corresponding authors}
\date{May 7, 2026}
\checkdata[Project Page]{\href{https://lkyyy111.github.io/VISD}{\color{headingblue}\texttt{https://lkyyy111.github.io/VISD}}}
\checkdata[GitHub]{\href{https://github.com/Koreyoshi01/VISD}{\color{headingblue}\texttt{https://github.com/Koreyoshi01/VISD}}}

\abstract{
  Training VideoLLMs for complex reasoning remains challenging due to sparse sequence-level rewards and the lack of fine-grained credit assignment over long, temporally grounded reasoning trajectories. While reinforcement learning with verifiable rewards (RLVR) provides reliable supervision, it fails to capture token-level contributions, leading to inefficient learning. Conversely, existing self-distillation methods offer dense supervision but lack structure and diagnostic specificity, and often interact unstably with reinforcement learning. In this work, we propose \textbf{VISD}, a structured self-distillation framework that introduces diagnostically meaningful privileged information for video reasoning. VISD employs a video-aware judge model to decompose reasoning quality into multiple dimensions, including answer correctness, logical consistency, and spatio-temporal grounding, and uses this structured feedback to guide a teacher policy for token-level supervision. To stably integrate dense supervision with RL, we adopt a direction--magnitude decoupling mechanism, where rollout-level advantages computed from rewards determine update direction, while structured privileged signals modulate token-level update magnitudes. This design enables semantically aligned and fine-grained credit assignment, improving both reasoning faithfulness and training efficiency. Additionally, VISD incorporates curriculum scheduling and EMA-based teacher stabilization to support robust optimization over long video sequences. Experiments on diverse benchmarks show that VISD consistently outperforms strong baselines, improving answer accuracy and spatio-temporal grounding quality. Notably, VISD reaches these gains with nearly $\boldsymbol{2\times}$ faster convergence in optimization steps, highlighting the effectiveness of structured self supervision in improving both performance and sample efficiency for VideoLLMs.
}

\begin{document}

\vspace*{-1.5cm}
\noindent
\begin{minipage}{\textwidth}
  \raisebox{0.115cm}{\includegraphics[height=0.82cm]{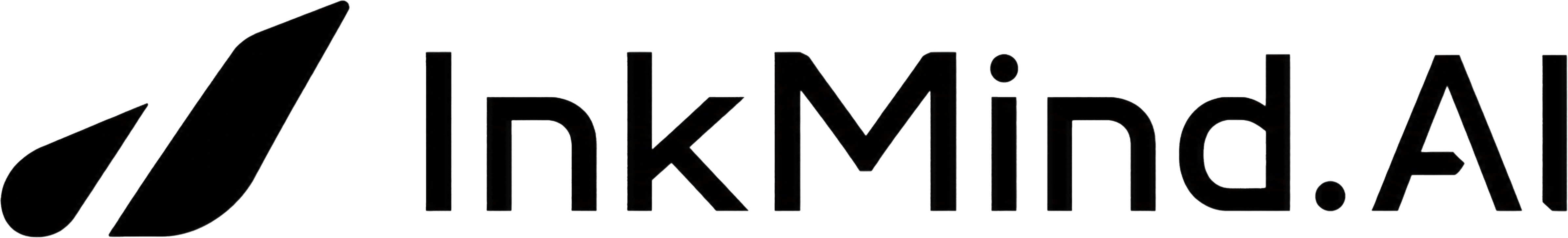}}
  \hfill
  \includegraphics[height=1.18cm]{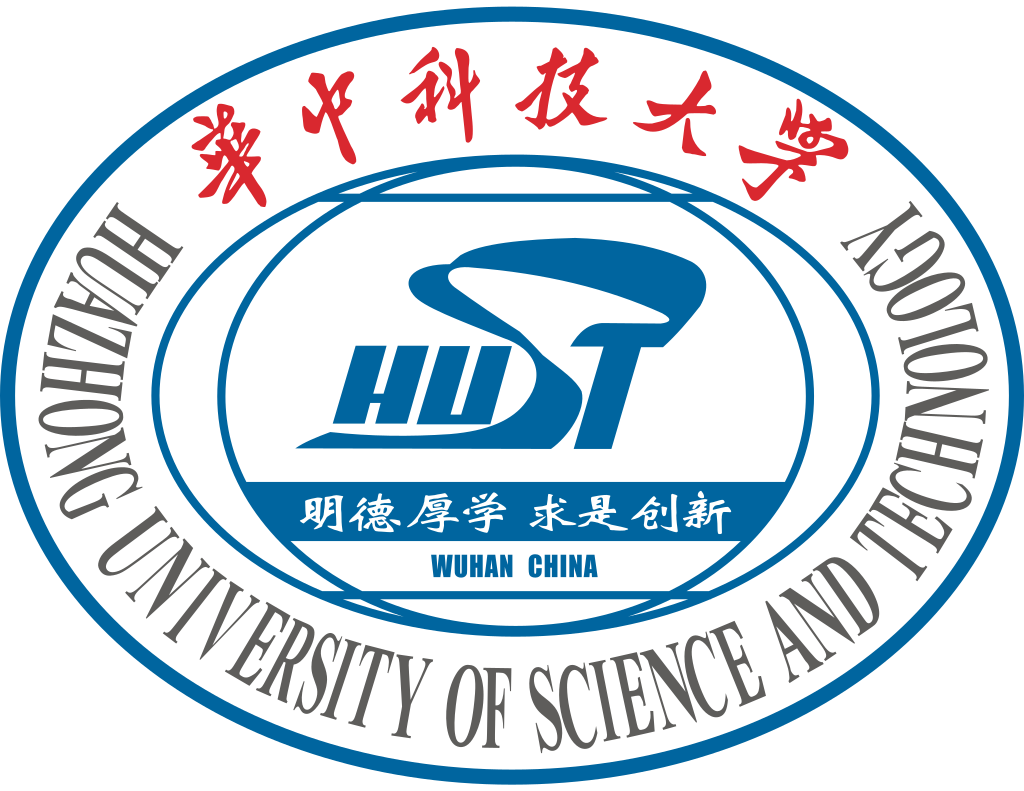}
  \hspace{0.18cm}
  \includegraphics[height=1.18cm]{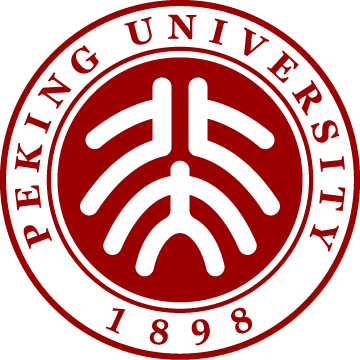}
  \hspace{0.18cm}
  \includegraphics[height=1.18cm]{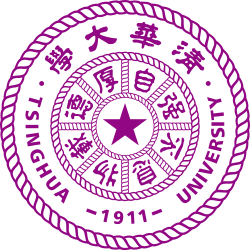}
  \hspace{0.18cm}
  \includegraphics[height=1.18cm]{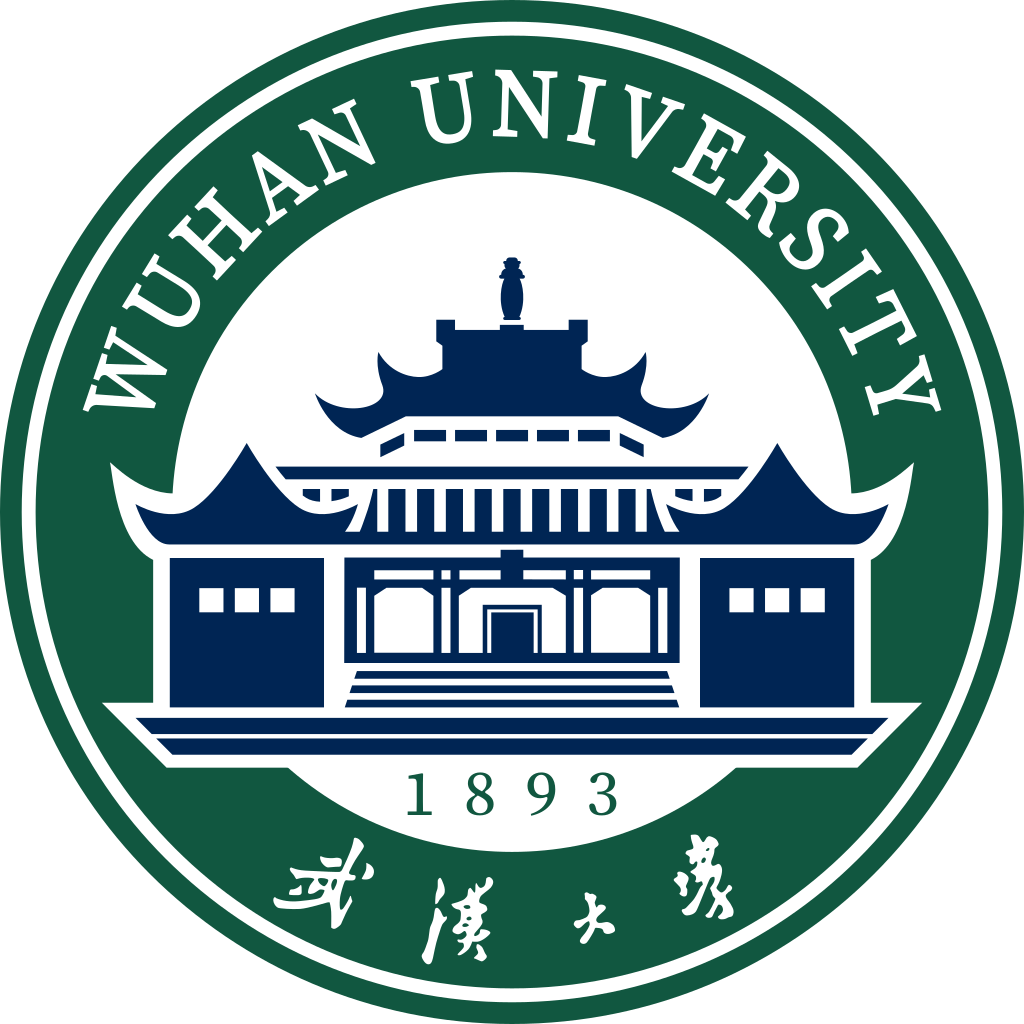}
\end{minipage}
\vspace{0.12cm}

\maketitle

\section{Introduction}
\label{sec:intro}

Recent progress in VideoLLMs~\cite{maaz2024videochatgpt,zhang2023videollama,fu2026videommev2stagebenchmarkscomprehensive,zeng2026video} has significantly advanced multimodal reasoning, enabling models to answer complex questions grounded in dynamic visual content.
Despite these improvements, training VideoLLMs remains fundamentally challenging. Unlike static image or text reasoning, video understanding requires aligning long reasoning trajectories with temporally evolving visual evidence~\cite{meng2026openo3video,feng2025videor1,wang2026video,sun2026mvp}, where both the correctness of the final answer and the faithfulness of intermediate reasoning are crucial.
This introduces a unique difficulty: learning signals must not only indicate whether an output is correct, but also identify \emph{where} and \emph{why} the reasoning process succeeds or fails.
Without such fine-grained guidance, training often becomes inefficient, requiring a large number of samples to propagate sparse rewards through long reasoning trajectories.

A natural approach to training such models is reinforcement learning with verifiable rewards~\cite{deepseekai2026deepseekr1,feng2025videor1,liao2026self} (RLVR), where correctness signals are derived from ground-truth answers or external verifiers. While effective in providing reliable optimization directions, these rewards are inherently sparse and operate at the sequence level~\cite{deepseekai2026deepseekr1,yang2026selfdistilledrlvr,zheng2025group}.
As a result, they fail to capture the heterogeneous contributions of individual tokens within long reasoning chains, especially when errors arise from subtle temporal misalignment or incorrect visual grounding. This limitation becomes more pronounced in video settings, where a correct answer may still be supported by flawed reasoning, and vice versa.

An alternative line of work leverages self-distillation to provide dense token-level supervision~\cite{zhao2026selfdistilledreasoner,hubotter2026reinforcementlearning,yang2026selfdistilledrlvr, shenfeld2601self}.
However, existing approaches typically treat the additional supervision as unstructured or modality-agnostic signals.
When directly applied to VideoLLMs, this leads to two key issues. First, the supervision lacks diagnostic specificity, making it difficult to distinguish between different types of reasoning errors, such as logical inconsistency versus grounding failure~\cite{ma2026fipo,peng2025simko}. Second, the interaction between self-distillation and reinforcement learning is often unstable, as auxiliary signals may override or conflict with reward-driven optimization~\cite{li2026unifying,li2026rethinking}.

In this work,
we propose \textbf{VISD}, a structured self-distillation framework that integrates reinforcement learning with structured privileged information. The central idea is to elevate self-distillation from a generic auxiliary signal to a structured supervision space that captures the compositional nature of reasoning errors in video tasks.
Concretely,
VISD introduces a video-aware judge model that evaluates each sampled reasoning trajectory and produces structured feedback along multiple dimensions, including answer correctness, reasoning consistency, and spatio-temporal grounding quality.
This feedback is used as privileged information to condition a teacher policy, which provides token-level guidance that is both fine-grained and diagnostically meaningful. In contrast to prior approaches~\cite{zhao2026selfdistilledreasoner,hubotter2026reinforcementlearning,yang2026selfdistilledrlvr,yan2025videochat} that rely on implicit or latent supervision, our design explicitly encodes the causes of reasoning failures, enabling more targeted corrections during training.

\begin{figure}[h]
  \vspace{-2mm}
  \centering
  \includegraphics[width=\linewidth]{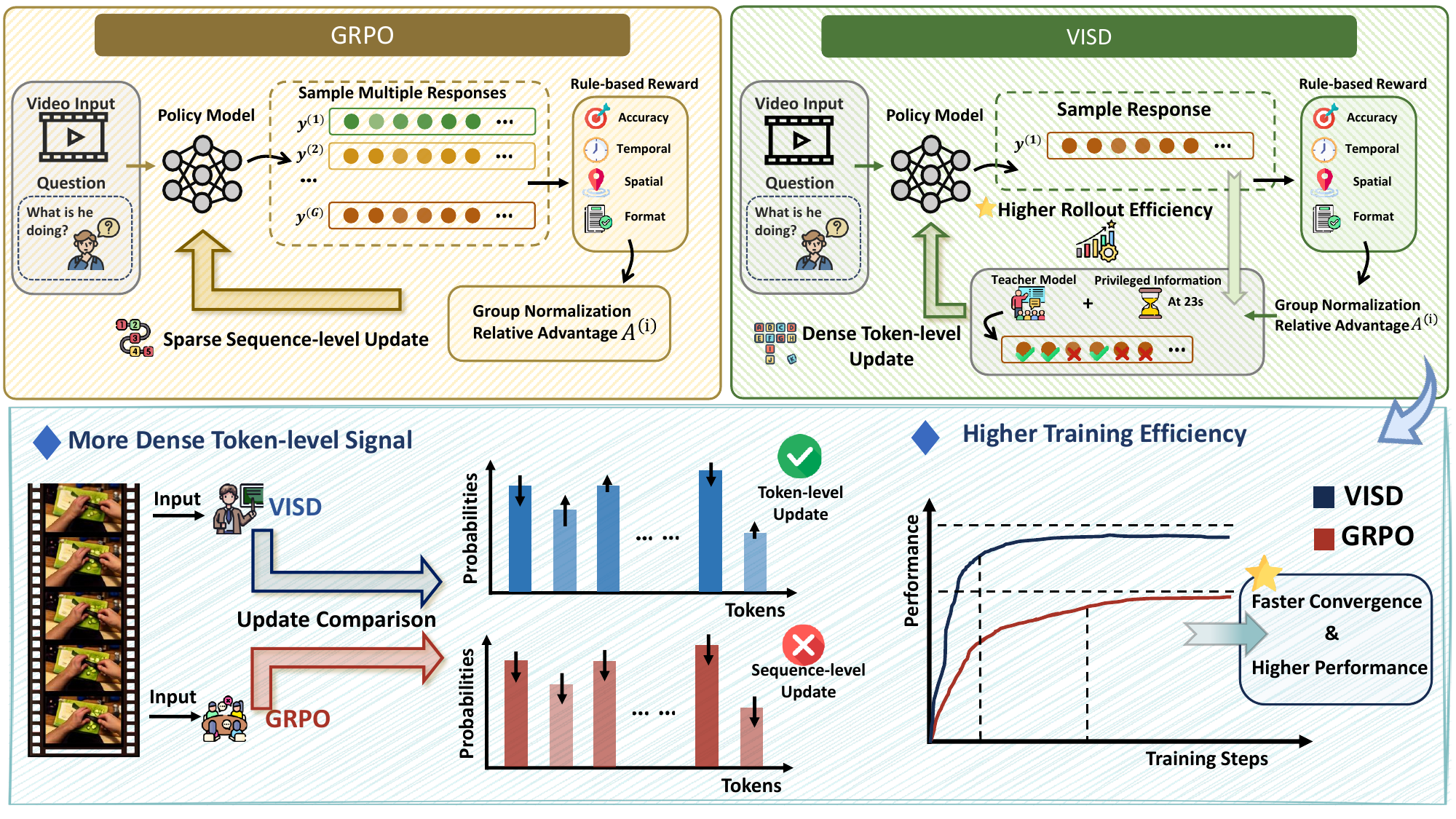}
  \vspace{-6mm}
  \caption{\emph{Comparison between RLVR method and VISD}. \textbf{(a)} The RLVR method, e.g., GRPO, provides sparse sequence-level signals and fails to capture fine-grained spatial and temporal evidence.
    \textbf{(b)} VISD uses dense token-level updates, thereby improving convergence speed and overall performance.}
  \label{fig:comparison}
  \vspace{-2mm}
\end{figure}

To effectively integrate structured supervision with reinforcement learning, we extend the direction--magnitude decoupling paradigm, with inspiration from RLSD~\cite{yang2026selfdistilledrlvr}. Specifically, rollout-level advantages computed from rewards determine the direction of policy updates, ensuring alignment with task objectives, while structured privileged information modulates the magnitude of token-level updates. This separation allows the model to benefit from dense supervision without compromising the stability of reward-driven optimization. Importantly, because the privileged information is structured, the resulting credit assignment is not only fine-grained but also semantically aligned with different aspects of video reasoning, as summarized in Figure~\ref{fig:comparison}.
As a result, VISD not only improves the quality and faithfulness of reasoning, but also significantly enhances learning efficiency by enabling faster convergence with more informative and structured training signals.

Training VideoLLMs further poses challenges due to long-horizon dependencies and heterogeneous reward signals~\cite{fu2026videommev2stagebenchmarkscomprehensive,meng2026openo3video,feng2025videor1}.
VISD addresses these issues through a set of principled optimization strategies. We adopt a curriculum that gradually transitions from structured self-distillation to reinforcement learning and maintain an exponential moving average teacher to stabilize token-level supervision. Together, these designs enable robust and scalable training under complex video reasoning scenarios.

We evaluate VISD on the Open-o3-Video benchmarks~\cite{meng2026openo3video} and demonstrate consistent improvements over strong baselines across a wide range of video reasoning tasks~\cite{fu2026videommev2stagebenchmarkscomprehensive,feng2025videor1}.
Beyond quantitative gains in accuracy, VISD also yields better grounding quality and more interpretable reasoning behaviors, highlighting the benefits of structured supervision.

In summary, our main contributions are as follows:
\begin{itemize}
  \item We propose \textbf{VISD}, a structured self-distillation framework tailored for video reasoning. VISD leverages a video-aware judge model to evaluate reasoning quality across multiple dimensions, providing diagnostically meaningful token-level supervision.
  \item We integrate a stable self-distillation training design tailored for video reasoning, combining direction--magnitude decoupled optimization with feedback-conditioned teacher replay, top-$K$ local support, curriculum annealing, and EMA teacher stabilization.
  \item We demonstrate the effectiveness of VISD through extensive evaluation on diverse video benchmarks. VISD consistently outperforms strong baselines, improving answer accuracy and spatio-temporal grounding quality while achieving nearly \textbf{2$\times$} faster convergence in optimization steps.
\end{itemize}

\section{Related Work}
\label{sec:related_work}

VISD is most closely related to three lines of work: video large language models, reinforcement learning with verifiable rewards, and on-policy/self-distillation methods for fine-grained credit assignment.
Recent VideoLLMs extend vision-language models to dynamic visual understanding through frame sampling, temporal aggregation, and video-specific instruction tuning~\cite{maaz2024videochatgpt,zhang2023videollama,zhang2025llavavideo,damonlpsg2025videollama3,fu2026videommev2stagebenchmarkscomprehensive,li2024mvbench,videommmu,cheng2025vstar}.
While these models improve video question answering and temporal reasoning, they are usually trained from static annotations and therefore provide limited supervision for identifying which step of a long reasoning trajectory causes an error.
RLVR methods address reasoning by optimizing sequence-level rewards~\cite{deepseekai2026deepseekr1,grpo,feng2025videor1,videochat-r1,wang2025videorft}, but their scalar advantages remain coarse for localized temporal, spatial, and object-grounding mistakes.
Self-distillation and on-policy distillation provide denser teacher-derived token signals~\cite{hinton2015distill,bornagain,policy_distill,agarwal2024policy,zhao2026selfdistilledreasoner,hubotter2026reinforcementlearning,yang2026selfdistilledrlvr}, yet existing formulations often treat the teacher signal as generic distributional guidance rather than an explicit diagnosis of multimodal reasoning failures.
VISD differs by conditioning the teacher on video-aware judge feedback: the rollout-level reward still determines the update direction, while feedback-conditioned teacher replay modulates token-level magnitudes according to answer consistency, temporal coverage, and spatial grounding errors.
Extended related works are provided in Appendix~\ref{app:extended_related_work}.




\section{Method}

\subsection{VISD Framework Overview}

We propose \textbf{VISD}, a structured self-distillation framework designed for video reasoning, where the core challenge lies in aligning long-horizon reasoning trajectories with fine-grained spatio-temporal evidence. Unlike conventional reinforcement learning approaches that rely on sparse sequence-level rewards~\citep{deepseekai2026deepseekr1,yang2026selfdistilledrlvr}, VISD introduces structured privileged supervision to enable token-level credit assignment that is both semantically meaningful and modality-aware.

Formally, given a video-question pair $x = (v, q)$, a policy model $\pi_\theta$ autoregressively generates a reasoning trajectory $y = (y_1, \dots, y_T)$. The trajectory may include intermediate reasoning steps, temporal references, object mentions, and the final answer. A verifier provides a scalar reward $R(x, y)$, which reflects task-level correctness and, when applicable, grounding quality. However, such reward signals are inherently sparse and fail to capture the heterogeneous contributions of individual tokens in long reasoning sequences.

To address this limitation, VISD augments the standard on-policy training pipeline with two auxiliary components. First, a teacher policy is instantiated from the same model but conditioned on privileged information, providing token-level assessments along the student's trajectory. Second, a dedicated judge model produces structured feedback tailored to video reasoning, which serves as the source of privileged information. The resulting framework retains the on-policy nature of reinforcement learning while introducing a structured signal pathway that refines credit assignment at token level.

\begin{figure}[t]
  \centering
  \includegraphics[width=\linewidth]{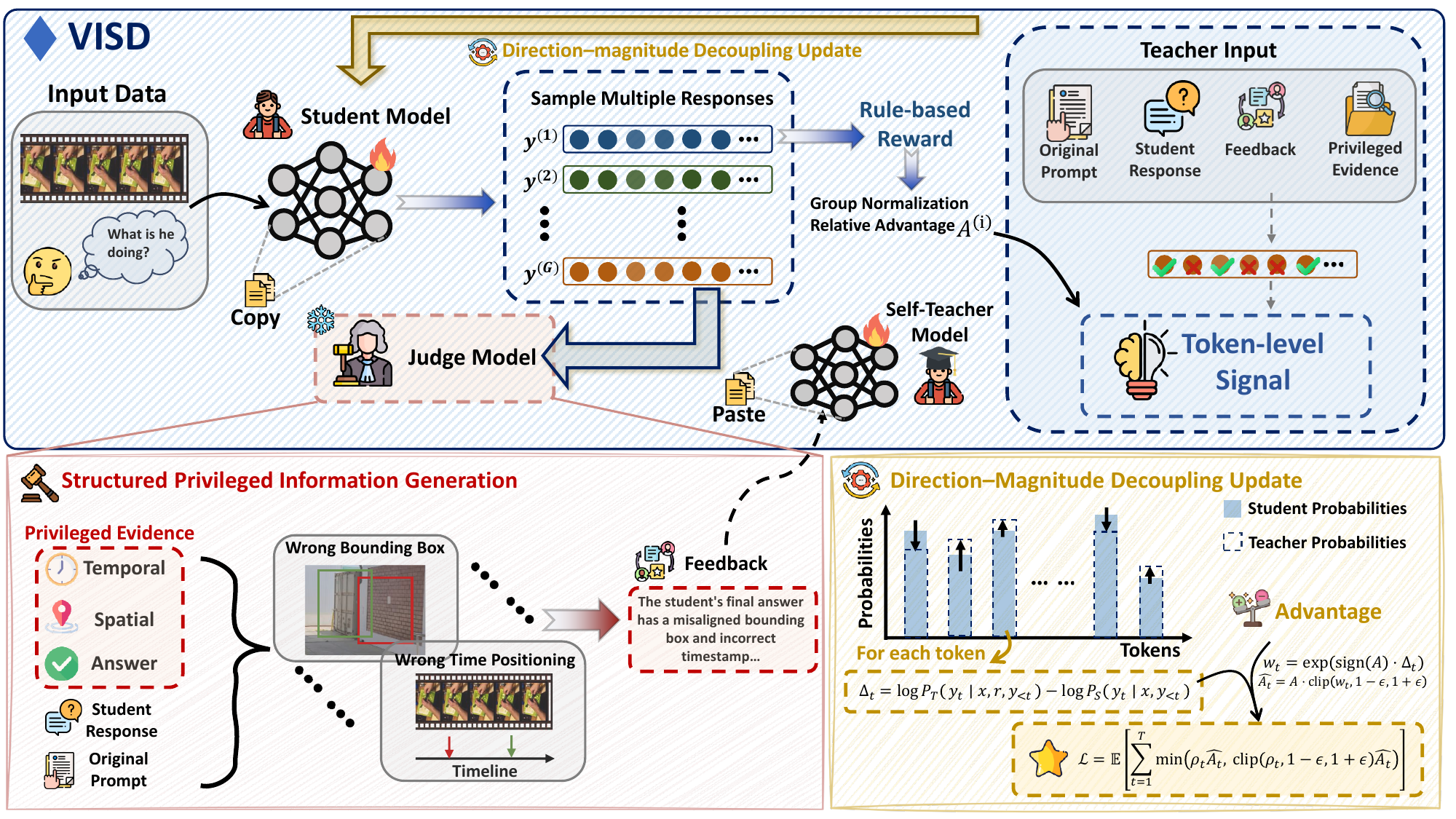}
  \vspace{-6mm}
  \caption{\small \emph{Overview of VISD.}
  VISD first uses a video-aware judge to generate structured privileged feedback for each student rollout, then replays the same completion with a feedback-conditioned teacher.
  The rollout-level reward advantage determines the update direction, while the teacher-student discrepancy modulates token-level update magnitudes for fine-grained credit assignment.}
  \label{fig:method}
  \vspace{-4mm}
\end{figure}

The overall training objective follows a policy-gradient formulation, illustrated in Figure~\ref{fig:method}:
\begin{equation}
    \nabla_\theta \mathcal{J} = \mathbb{E}_{y \sim \pi_\theta(\cdot|x)} \left[ \sum_{t=1}^{T} \hat{A}_t \nabla_\theta \log \pi_\theta(y_t | x, y_{<t}) \right]
    \label{eq: policy gradient}
\end{equation}
where $\hat{A}_t$ denotes a token-level advantage that integrates both environment reward and structured privileged information. Crucially, VISD preserves the reliability of environment-driven optimization while enriching it with fine-grained, structured signals that are particularly suited for video reasoning.

The full VISD training procedure is summarized in Algorithm~\ref{alg:visd} in Appendix~\ref{app:algorithm_details}.

\subsection{Structured Privileged Information Generation}

A key observation underlying our design is that, in video reasoning, not all errors are equal: incorrect predictions may arise from temporal misalignment, spatial grounding failures, or inconsistencies between reasoning and answer. Capturing such distinctions is essential for effective learning, yet existing self-distillation methods~\citep{zhao2026selfdistilledreasoner,hubotter2026reinforcementlearning,yang2026selfdistilledrlvr} treat privileged information as an unstructured signal, limiting their ability to guide fine-grained corrections.

To overcome this limitation, VISD introduces a judge model $J$ that transforms each rollout trajectory into a structured representation of reasoning quality. Given an input $x$, a sampled trajectory $y$, verified answer-side information $a^\star$, and available grounding evidence $e$, the judge produces structured feedback $f = J(x,y,a^\star,e)$. We package the teacher-side privileged context as $r=(a^\star,e,f)$. The feedback $f$ encapsulates multiple complementary aspects of evaluation. Specifically, it assesses (i) the correctness of the predicted answer, (ii) the consistency between the reasoning process and the answer, and (iii) the underlying causes of inconsistencies when they arise. More importantly, to account for the unique challenges of video understanding, the judge incorporates modality-specific feedback by explicitly evaluating temporal alignment and spatial grounding.
This includes identifying whether the reasoning references incorrect time windows, whether objects are mislocalized, and whether the cited evidence is inconsistent with the visual content, etc.
The teacher distribution is thus defined as:
\begin{equation}
    P_T(y_t | x, r, y_{<t}) = \pi_{\bar{\theta}}(y_t | x, r, y_{<t})
    \label{eq: teacher policy}
\end{equation}
where $\bar{\theta}$ denotes the teacher parameters; in the main EMA setting, $\bar{\theta}\equiv\theta_{\mathrm{ema}}$.
The conditioning on $r$ allows the teacher to rescore the same student completion using verified answer/evidence and rollout-specific judge feedback.

\subsection{Direction--Magnitude Decoupled Structured Self-Distillation}

Given a sampled trajectory, we compute the token-level discrepancy between teacher and student distributions on a compact top-$K$ local support.
At each position $t$, we construct a teacher-centered support
\begin{equation}
    \mathcal{U}_t
    =
    \operatorname{TopK}_{P_T(\cdot \mid x,r,y_{<t})}(K)
    \cup
    \{y_t\},
    \label{eq:topk_support_visd}
\end{equation}
which augments the teacher top-$K$ candidates with the realized on-policy token so that the generated action is always covered.
Let $\widetilde P_T(\cdot \mid x,r,y_{<t})$ and $\widetilde P_S(\cdot \mid x,y_{<t})$ denote the teacher and student distributions renormalized over $\mathcal{U}_t$.
We define:
\begin{equation}
    \Delta_t =
    \mathrm{sg}\left(
    \log \widetilde P_T(y_t | x, r, y_{<t})
    -
    \log \widetilde P_S(y_t | x, y_{<t})
    \right)
    \label{eq: discrepancy}
\end{equation}
where $\mathrm{sg}(\cdot)$ denotes the stop-gradient operator.
This top-$K$ local-support discrepancy measures how the structured privileged information revises the model's belief about the realized token within a locally relevant candidate set.
In contrast to standard self-distillation, where such discrepancies are treated as flat signals, here $\Delta_t$ implicitly encodes structured corrections derived from the judge, including temporal, spatial, and logical inconsistencies.
We provide the detailed top-$K$ support derivation, including direction preservation and bounded magnitude redistribution, in Appendix~\ref{app:topk_local_support_ratio}.

Unlike existing mainstream self-distillation paradigms,
VISD adopts a direction--magnitude decoupling strategy to integrate structured privileged information into policy optimization. The central principle is to separate \emph{what direction the policy should move} from \emph{how strongly each token should be updated}, thereby preventing unreliable signals from corrupting the optimization trajectory.

The sequence-level advantage $A$, computed from the reward, determines whether a trajectory should be reinforced or penalized. We then define a token-level reweighting factor:
\begin{equation}
    w_t = \exp\big( \mathrm{sign}(A) \cdot \Delta_t \big)
    \label{eq: weight}
\end{equation}

When the trajectory has positive advantage, tokens favored by the teacher receive larger update magnitudes; when it has negative advantage, tokens disfavored by the teacher receive stronger suppression. To ensure stability, we clip the reweighting factor:
\begin{equation}
    m_t = \mathrm{clip}(w_t, 1-\epsilon_w, 1+\epsilon_w),
    \label{eq: advantage}
\end{equation}
where $\epsilon_w$ bounds the amount of token-level credit redistribution.
This formulation keeps the sign of the token-level advantage determined by the rollout-level advantage, while the structured privileged information redistributes credit within the trajectory.
Because the teacher is conditioned on diagnostically meaningful feedback, the resulting reweighting reflects not only confidence differences, but also the semantic causes of errors.

\subsection{Stable Policy Optimization for Long Video Sequences}

Optimizing VideoLLMs over long reasoning trajectories is inherently challenging due to sparse rewards, heterogeneous supervision signals, and the instability introduced by token-level credit assignment. In practice, these challenges are further amplified by the multi-dimensional nature of video reasoning, where correctness, reasoning consistency, and spatio-temporal grounding often induce rewards with different scales and statistical properties. VISD addresses these issues through a set of design choices that ensure stable and robust optimization.


We adopt a \emph{curriculum strategy} that gradually transitions from structured self-distillation to reinforcement learning. At the early stage of training, updates are modulated by the teacher-induced token-level reweighting, allowing the model to quickly acquire fine-grained reasoning patterns guided by structured privileged information. As training progresses, the influence of self-distillation is gradually annealed, and the optimization increasingly relies on environment rewards. Formally, for rollout $i$, the final token-level advantage is defined as:
\begin{equation}
    \hat{A}^{(i)}_t = A^{(i)} \cdot \big( (1 - \lambda) + \lambda \cdot m^{(i)}_t \big)
    \label{eq: lambda}
\end{equation}
where $\lambda \in [0,1]$ controls the strength of teacher-induced token reweighting.
This formulation effectively interpolates between structured self-distillation and standard reinforcement learning, ensuring both efficient early learning and stable long-term convergence.

To stabilize the teacher signal, we maintain an exponential moving average version of the policy parameters to construct the teacher model. Specifically, the teacher parameters $\bar{\theta}$ are updated as:
\begin{equation}
    \bar{\theta} \leftarrow \tau \bar{\theta} + (1 - \tau)\theta
    \label{eq: ema}
\end{equation}
where $\tau$ is a decay factor close to 1.
The teacher distribution is then defined using $\pi_{\bar{\theta}}$ instead of the instantaneous student parameters.
This design reduces high-frequency fluctuations in the teacher signal and mitigates instability caused by rapidly changing policy distributions, which is especially critical in long-horizon video reasoning.

Finally, we optimize the policy using the GRPO-style clipped surrogate objective:
\begin{equation}
    \begin{aligned}
    \mathcal{J}_{\mathrm{VISD}}
    =
    \mathbb{E} \left[
    \frac{1}{G}\sum_{i=1}^{G}
    \frac{1}{|y^{(i)}|}\sum_{t=1}^{|y^{(i)}|}
    \min\left(
    \rho^{(i)}_t \hat{A}^{(i)}_t,
    \mathrm{clip}(\rho^{(i)}_t, 1-\epsilon_{\mathrm{pg}}, 1+\epsilon_{\mathrm{pg}})\hat{A}^{(i)}_t
    \right)
    \right]
    \end{aligned}
    \label{eq: obj}
\end{equation}
where $\rho^{(i)}_t$ is the importance sampling ratio between the current and old policies for token $t$ in rollout $i$, and $\epsilon_{\mathrm{pg}}$ is the policy-gradient clipping range.
We do not add a reference-model KL penalty or a separate distillation loss; the teacher only modifies the magnitude of the GRPO advantage.
This preserves the stability benefits of trust-region optimization while incorporating structured token-level credit assignment.
These mechanisms together enable VISD to effectively scale structured self-distillation to long video sequences, achieving stable optimization despite the complexity of multi-dimensional rewards and long-horizon dependencies.


\section{Experiments}

\subsection{Experimental Setup}

\paragraph{\normalfont\textbf{Benchmarks.}}
We evaluate VISD on the Open-o3-Video~\cite{meng2026openo3video} benchmark suite, which consists of diverse video reasoning tasks covering temporal understanding, object grounding, and compositional reasoning.
The benchmarks include both short and long video scenarios, requiring models to reason over dynamic visual content and produce multi-step explanations.
We also evaluate on Video-MME-v2~\cite{fu2026videommev2stagebenchmarkscomprehensive}, which provides a more challenging video understanding setting with fine-grained level-wise and consistency-oriented metrics.
We additionally evaluate zero-shot temporal grounding on Charades-STA~\cite{gao2017tall}.

\paragraph{\normalfont\textbf{Implementation Details.}}
VISD starts from the same SFT initialization recipe as Open-o3-Video and then applies our self-distillation RL training for video reasoning.
The SFT initialization establishes the grounded reasoning format, while the VISD RL stage optimizes answer correctness and grounding quality on STGR-RL prompts through feedback-conditioned teacher replay.
During RL, VISD uses feedback-conditioned teacher replay only as a training-time token-credit signal.
Unless otherwise specified, all VISD (Ours) results are evaluated with the student checkpoint at optimization step 1200.
The Open-o3-Video and VisionCoach baselines are evaluated using their fully trained RL checkpoints, which require more than 2300 optimization steps; thus, VISD reaches stronger performance with about 50\% of the optimization steps.
More implementation, training, and evaluation details are provided in Appendices~\ref{app:algorithm_details}, \ref{app:training_details}, and~\ref{app:evaluation_details}.

\subsection{Main Results}

\begin{table*}[t]
  \vspace{-4mm}
  \centering
\caption{\small \emph{V-STAR results.}
  Chain1 denotes \emph{what--when--where}, while Chain2 denotes \emph{what--where--when}. $^*$ indicates rows evaluated by us.}
  \resizebox{0.99\textwidth}{!}{
    \begin{tabular}{lccccccc}
      \toprule[0.15em]
      \textbf{Model}                                   &
      \textbf{What}                                    &
      \multicolumn{2}{c}{\textbf{When (Temporal IoU)}} &
      \multicolumn{2}{c}{\textbf{Where (Spatial IoU)}} &
      \multicolumn{2}{c}{\textbf{Overall}}                                                                                                                            \\
      \cmidrule(lr){2-2} \cmidrule(lr){3-4} \cmidrule(lr){5-6} \cmidrule(lr){7-8}
                                                       &
      \textbf{Acc}                                     &
      \textbf{Chain1}                                  &
      \textbf{Chain2}                                  &
      \textbf{Chain1}                                  &
      \textbf{Chain2}                                  &
      \textbf{mAM}                                     &
      \textbf{mLGM}                                                                                                                                                   \\
      \midrule
      \rowcolor{red!8} \multicolumn{8}{l}{\emph{Proprietary Models}}                                                                                                  \\
      Gemini-2-Flash                                   & 53.0          & 24.5          & 23.8          & 4.6           & 2.2          & 26.9          & 35.6          \\
      GPT-4o                                           & 60.8          & 16.7          & 12.8          & 6.5           & 3.0          & 26.8          & 38.2          \\
      \midrule
      \rowcolor{green!8} \multicolumn{8}{l}{\emph{Open-source Models}}                                                                                                \\
      TRACE                                            & 17.6          & 19.1          & 17.1          & 0.0           & 0.0          & 12.0          & 13.3          \\
      Oryx-1.5-7B                                      & 20.5          & 13.5          & 14.8          & 10.1          & 3.5          & 15.1          & 13.8          \\
      VideoChat2                                       & 36.2          & 13.7          & 12.5          & 2.5           & 1.0          & 17.0          & 20.3          \\
      Qwen2.5-VL-7B$^*$                                & 33.5          & 15.4          & 13.8          & 17.0          & 2.5          & 19.3          & 22.4          \\
      InternVL-2.5-8B                                  & 44.2          & 8.7           & 7.8           & 0.7           & 0.1          & 17.6          & 24.9          \\
      Video-LLaMA3                                     & 41.9          & 23.0          & 23.1          & 0.9           & 0.2          & 21.7          & 27.0          \\
      LLaVA-Video                                      & 49.5          & 10.5          & 12.2          & 1.9           & 1.3          & 20.8          & 27.3          \\
      Open-o3-Video-7B$^*$                             & 60.1          & 25.0          & 23.6          & 25.6          & 5.6          & 33.4          & 45.8          \\
      VisionCoach-7B$^*$                               & 61.2          & 25.4          & 25.4          & \textbf{27.1} & 5.3          & 34.3          & 47.5          \\
      \midrule
      \textbf{VISD (Ours)}                                      & \textbf{61.9} & \textbf{26.8} & \textbf{27.2} & 26.7          & \textbf{6.3} & \textbf{35.1} & \textbf{48.9} \\
      $\Delta$ vs.\ Qwen2.5-VL-7B                      & +28.4         & +11.4         & +13.4         & +9.7          & +3.8         & +15.8         & +26.5         \\
      \bottomrule[0.15em]
    \end{tabular}
  }
  \label{tab:vstar}
  \vspace{-2mm}
\end{table*}

\begin{table*}[t]
  \centering
\caption{\small \emph{General benchmark results.}
  ``LRR'' refers to LongVideo-Reason-eval~\citep{longvila-r1}. The Average score is computed over the bold-faced dataset-level metrics. $^*$ indicates rows evaluated by us; for rows marked with $^\dagger$, the WorldSense scores are evaluated by us.}
  \resizebox{0.99\textwidth}{!}{
    \begin{tabular}{lccccccccccc}
      \toprule[0.15em]
      \textbf{Model}                          &
      \multicolumn{4}{c}{\textbf{WorldSense}} &
      \multicolumn{4}{c}{\textbf{VideoMMMU}}  &
      \textbf{LRR}                            &
      \textbf{TVGBench}                       &
      \textbf{Average} \\
      \cmidrule(lr){2-5} \cmidrule(lr){6-9} \cmidrule(lr){10-10} \cmidrule(lr){11-11}
                                              & \textbf{Overall} & Recognition   & Understanding  & Reasoning
                                              & \textbf{Overall} & Comprehension & Adaptation     & Perception
                                              & \textbf{Acc}     & \textbf{mIoU} & \textbf{Score} \\
      \midrule
      \rowcolor{red!8} \multicolumn{12}{l}{\emph{Proprietary Models}} \\
      GPT-4o                                  & 42.6             & -             & -              & -             & 61.2          & -             & -             & 66.0          & -    & -             & -    \\
      \midrule
      \rowcolor{blue!8} \multicolumn{12}{l}{\emph{Tool-calling Methods}} \\
      \rowcolor{gray!8} LongVT-RL-7B$^*$      & \color{gray} 36.1 & \color{gray} 33.1 & \color{gray} 39.1 & \color{gray} 37.8 & \color{gray} 52.1 & \color{gray} 49.3 & \color{gray} 43.0 & \color{gray} 64.0 & \color{gray} 75.1 & \color{gray} 10.6 & \color{gray} 43.5 \\
      \midrule
      \rowcolor{green!8} \multicolumn{12}{l}{\emph{Tool-free Methods}} \\
      Qwen2.5-VL-7B$^\dagger$                 & 35.5             & 32.9          & 38.4           & 37.0          & 51.2          & -             & -             & 64.7          & 59.3 & 16.3          & 40.6 \\
      VideoRFT-7B$^\dagger$                   & 38.4             & 36.5          & 40.0           & 40.1          & 51.1          & -             & -             & 66.0          & 69.4 & 14.3          & 43.3 \\
      VideoR1-7B$^\dagger$                    & 36.9             & 35.3          & 37.7           & 39.0          & 52.4          & -             & -             & 65.3          & 68.9 & 9.6           & 42.0 \\
      Open-o3-Video-7B$^*$                    & 38.9             & 37.4          & 39.7           & 40.5          & 52.6          & 49.1          & 41.3          & 67.4          & 69.1 & 19.2          & 44.9 \\
      VisionCoach-7B$^*$                      & 39.2             & 38.4          & 40.2           & 39.6          & \textbf{53.8} & 49.0          & 42.2          & \textbf{70.3} & 70.7 & 21.0          & 46.2 \\
      \midrule
      \textbf{VISD (Ours)}                   & \textbf{39.9}    & \textbf{39.4} & \textbf{40.4} & \textbf{40.5} & \textbf{53.8} & \textbf{49.7} & \textbf{42.3} & 69.3          & \textbf{73.5} & \textbf{24.8} & \textbf{48.0} \\
      $\Delta$ vs.\ Qwen2.5-VL-7B             & +4.4             & +6.5          & +2.0           & +3.5          & +2.6          & -             & -             & +4.6          & +14.2 & +8.5          & +7.4 \\
      \bottomrule[0.15em]
    \end{tabular}
  }
  \label{tab:general_video_understanding}
\end{table*}

\begin{table*}[t]
  \centering
\caption{\small \emph{Video-MME-v2 results.}
  We report the official non-linear score, level-wise scores, consistency, coherence, and average accuracy using 64 sampled frames without subtitles.}
  \resizebox{0.99\textwidth}{!}{
    \begin{tabular}{lcccccccc}
      \toprule[0.15em]
      \textbf{Model}      & \textbf{Frames} & \textbf{Non-Lin Score} & \textbf{Level 1} & \textbf{Level 2} & \textbf{Level 3} & \textbf{Consistency} & \textbf{Coherence} & \textbf{Avg Acc} \\
      \midrule
      Qwen2.5-VL-7B       & 64              & 7.9                    & 11.4             & 6.7              & 6.7              & 8.2                  & 7.3                & 21.3             \\
      VideoRFT-7B         & 64              & 8.2                    & 11.8             & 7.4              & 6.8              & 8.8                  & 7.2                & 21.7             \\
      Video-R1-7B         & 64              & 7.7                    & 11.1             & 6.7              & 6.4              & 8.0                  & 7.2                & 20.7             \\
      Open-o3-Video-7B    & 64              & 7.8                    & 11.0             & 6.9              & 6.6              & 8.3                  & 6.9                & 21.2             \\
      VisionCoach-7B      & 64              & 8.6                    & 12.1             & 7.3              & 7.5              & 9.0                  & 7.9                & 22.1             \\
      \textbf{VISD (Ours)} & 64              & \textbf{9.4}           & \textbf{13.4}    & \textbf{8.2}     & \textbf{7.9}     & \textbf{10.2}        & \textbf{8.0}       & \textbf{23.8}    \\
      \bottomrule[0.15em]
    \end{tabular}
  }
  \label{tab:videommev2_results}
\end{table*}

\paragraph{\normalfont\textbf{Spatio-Temporal Grounding on V-STAR}}~\cite{cheng2025vstar}.
Table~\ref{tab:vstar} compares VISD with closed-source models~\cite{gpt4o,comanici2025gemini}, general video LLMs~\cite{qwen2.5vl}, and recent reasoning-focused frameworks~\cite{feng2025videor1,meng2026openo3video,lee2026visioncoach}.
VISD improves answer accuracy over Qwen2.5-VL-7B by +28.4 points and obtains the best overall V-STAR scores, improving mAM/mLGM over VisionCoach from 34.3/47.5 to 35.1/48.9.
The gains appear not only in final answering, but also in temporal localization and spatial grounding, which is the setting where sparse sequence-level rewards are least informative.

\newpage
\begin{wraptable}{r}{0.45\textwidth}
  \centering
\caption{\small \emph{Charades-STA results.}
  We report zero-shot temporal grounding with recall at IoU thresholds 0.3, 0.5, and 0.7, together with mIoU.}
  \setlength{\tabcolsep}{3.2pt}
  \renewcommand{\arraystretch}{0.9}
  \resizebox{\linewidth}{!}{
    \begin{tabular}{lcccc}
      \toprule[0.1em]
      \textbf{Model} & \textbf{R@0.3} & \textbf{R@0.5} & \textbf{R@0.7} & \textbf{mIoU} \\
      \midrule
      \rowcolor{green!8}\multicolumn{5}{l}{\emph{General Video LLMs}}                   \\
      VideoChat      & 9.0            & 3.3            & 1.3            & 6.5           \\
      VideoLLaMA     & 10.4           & 3.8            & 0.9            & 7.1           \\
      Video-ChatGPT  & 20.0           & 7.7            & 1.7            & 13.7          \\
      Valley         & 28.4           & 1.8            & 0.3            & 21.4          \\
      \rowcolor{green!8}\multicolumn{5}{l}{\emph{Temporal Grounding Video LLMs}}        \\
      GroundingGPT   & -              & 29.6           & 11.9           & -             \\
      Momentor       & 42.6           & 26.6           & 11.6           & 28.5          \\
      TimeChat       & -              & 32.2           & 13.4           & -             \\
      HawkEye        & 50.6           & 31.4           & 14.5           & 33.7          \\
      VTG-LLM        & 51.2           & 33.8           & 15.7           & 34.4          \\
      VTimeLLM       & 51.0           & 27.5           & 11.4           & 31.2          \\
      \rowcolor{green!8}\multicolumn{5}{l}{\emph{Grounded Reasoning Models}}            \\
      Open-o3-video  & 62.6           & 45.6           & 24.5           & 42.5          \\
      VisionCoach    & 63.2           & 45.8           & \textbf{24.7}  & 42.7          \\
      \textbf{VISD (Ours)}    & \textbf{67.3}  & \textbf{46.1}  & 21.6           & \textbf{44.6} \\
      \bottomrule[0.1em]
    \end{tabular}
  }
  \label{tab:charades_sta}
\end{wraptable}

\paragraph{\normalfont\textbf{General Video Understanding.}}
Beyond V-STAR, VISD also improves the average score across broader video reasoning benchmarks~\cite{fu2026videommev2stagebenchmarkscomprehensive,videommmu,worldsense,longvila-r1,timer1}.
As shown in Table~\ref{tab:general_video_understanding}, VISD outperforms Qwen2.5-VL-7B by +7.4 average points and improves over VisionCoach by +1.8 average points, with the strongest gains on long-video reasoning and temporal grounding metrics such as LRR and TVGBench.
On Video-MME-v2, VISD also improves the non-linear score from 8.6 to 9.4 over VisionCoach and obtains the best average accuracy (Table~\ref{tab:videommev2_results}).
These results suggest that feedback-conditioned token supervision preserves general video QA ability while improving the grounding-oriented behaviors targeted by VISD.
Additional qualitative examples are provided in Appendix~\ref{app:qualitative_case_studies}.

\paragraph{\normalfont\textbf{Temporal Grounding on Charades-STA.}}
On Charades-STA~\cite{gao2017tall}, VISD improves R@0.3, R@0.5, and mIoU over Open-o3-Video and VisionCoach (Table~\ref{tab:charades_sta}), indicating stronger zero-shot moment localization.
This is a useful stress test because the model must output a temporal segment without task-specific fine-tuning on the target benchmark.

\subsection{Ablation Studies}
\label{sec:ablation}

We study three questions that correspond to the main design choices in VISD.
First, does structured judge feedback provide additional benefit beyond using verified answer information alone?
Second, does the improvement come from a particular teacher-parameter update heuristic, or is the update rule mainly a stabilization choice?
Third, is it sufficient to compare the teacher only on the realized sampled token, or does VISD's local top-$K$ support provide a more useful token-level magnitude signal?
We pair reward curves with downstream benchmark tables because training reward alone may not fully reflect grounding quality.

\begin{figure}[t]
  \vspace{-2mm}
  \centering
  \begin{minipage}[t]{0.32\linewidth}
    \centering
    \includegraphics[width=\linewidth]{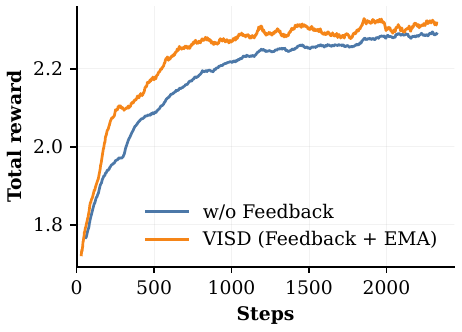}\\[-0.2em]
    {\footnotesize\textbf{(a)}}
  \end{minipage}
  \hfill
  \begin{minipage}[t]{0.32\linewidth}
    \centering
    \includegraphics[width=\linewidth]{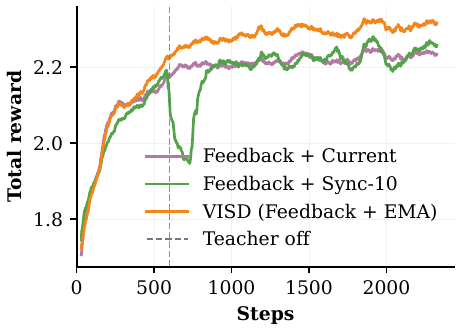}\\[-0.2em]
    {\footnotesize\textbf{(b)}}
  \end{minipage}
  \hfill
  \begin{minipage}[t]{0.32\linewidth}
    \centering
    \includegraphics[width=\linewidth]{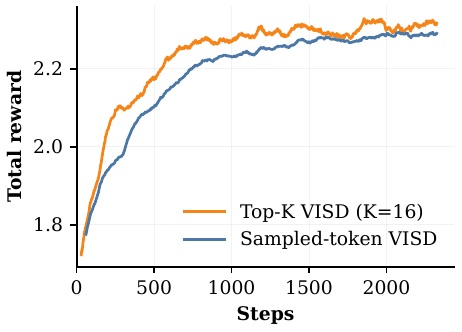}\\[-0.2em]
    {\footnotesize\textbf{(c)}}
  \end{minipage}
  \vspace{-2mm}
  \caption{\small \emph{Training ablation curves.}
  \textbf{(a)} Feedback conditioning compares VISD with and without judge feedback, showing that structured feedback leads to a stronger final reward trajectory.
  \textbf{(b)} Teacher update strategy compares current-policy, Sync-10, and EMA teachers, where EMA provides the most stable optimization and best final reward.
  \textbf{(c)} Top-$K$ support versus sampled token compares two teacher-student credit formulations, showing a modest but consistent advantage for VISD's local top-$K$ support.}
  \label{fig:training_ablation_curves}
  \vspace{-2mm}
\end{figure}

Table~\ref{tab:training_ablation} reports the final training statistics corresponding to the ablation curves in Figure~\ref{fig:training_ablation_curves}.
Reward and answer accuracy summarize final optimization quality, while gradient norm serves as a stability diagnostic for teacher-update strategies.

\begin{table}[!htbp]
  \vspace{-2mm}
  \centering
  \small
  \caption{\small \emph{Training ablation on feedback conditioning and teacher update strategy.}
  We report final training reward, answer accuracy, and gradient norm for the main feedback and teacher-parameter variants, averaged over the final 100 training steps.}
  \label{tab:training_ablation}
  \begin{tabular}{lccccc}
    \toprule
    Method & Feedback & Teacher & Reward $\uparrow$ & Ans. Acc. $\uparrow$ & Grad Norm $\downarrow$ \\
    \midrule
    VISD w/o Feedback & No & EMA & 2.280 & 0.565 & \textbf{2.63} \\
    VISD w/ current-policy teacher & Yes & Current & 2.228 & 0.557 & 3.49 \\
    VISD w/ Sync-10 teacher & Yes & Sync-10 & 2.253 & 0.558 & 11.63 \\
    \textbf{VISD w/ EMA teacher (Ours)} & Yes & EMA & \textbf{2.318} & \textbf{0.571} & 3.58 \\
    \bottomrule
  \end{tabular}
  \vspace{-1mm}
\end{table}

\begin{table}[t]
  \vspace{-1mm}
  \centering
  \small
  \caption{\small \emph{VSTAR ablation results.}
  We compare feedback/teacher variants and top-$K$ versus sampled-token credit on answer accuracy, temporal IoU, spatial IoU, and overall VSTAR metrics.}
  \label{tab:vstar_ablation}
  \resizebox{\linewidth}{!}{
    \begin{tabular}{lccccccc}
      \toprule[0.15em]
      \textbf{Model} &
      \textbf{What} &
      \multicolumn{2}{c}{\textbf{When (Temporal IoU)}} &
      \multicolumn{2}{c}{\textbf{Where (Spatial IoU)}} &
      \multicolumn{2}{c}{\textbf{Overall}} \\
      \cmidrule(lr){2-2} \cmidrule(lr){3-4} \cmidrule(lr){5-6} \cmidrule(lr){7-8}
      &
      \textbf{Acc} &
      \textbf{Chain1} &
      \textbf{Chain2} &
      \textbf{Chain1} &
      \textbf{Chain2} &
      \textbf{mAM} &
      \textbf{mLGM} \\
      \midrule
      \rowcolor{green!8} \multicolumn{8}{l}{\emph{Ablation Variants}} \\
      VISD w/o Feedback & 59.7 & 25.2 & 24.8 & 23.7 & 5.5 & 33.1 & 45.3 \\
      VISD w/ current-policy teacher & 60.2 & 23.8 & 21.7 & 23.5 & 5.7 & 32.5 & 44.8 \\
      VISD w/ Sync-10 teacher & 58.6 & \textbf{28.8} & \textbf{27.3} & 11.2 & 2.7 & 31.2 & 42.8 \\
      \textbf{VISD w/ EMA teacher (Ours)} & \textbf{61.9} & 26.8 & 27.2 & \textbf{26.7} & \textbf{6.3} & \textbf{35.1} & \textbf{48.9} \\
      \midrule
      \rowcolor{green!8} \multicolumn{8}{l}{\emph{Top-$K$ Analysis}} \\
      VISD w/ sampled token & 61.3 & 25.2 & 25.4 & 24.8 & 6.1 & 34.0 & 47.0 \\
      \textbf{VISD w/ top-$K$ (Ours)} & \textbf{61.9} & \textbf{26.8} & \textbf{27.2} & \textbf{26.7} & \textbf{6.3} & \textbf{35.1} & \textbf{48.9} \\
      \bottomrule[0.15em]
    \end{tabular}
  }
  \vspace{-1mm}
\end{table}

\begin{table}[!htbp]
  \vspace{-1mm}
  \centering
  \small
  \caption{\small \emph{General video reasoning ablation results.}
  We evaluate the same ablation variants on broad video understanding and temporal-grounding benchmarks, reporting the average over dataset-level metrics.}
  \label{tab:general_ablation}
  \resizebox{0.99\textwidth}{!}{
    \begin{tabular}{lccccccccccc}
      \toprule[0.15em]
      \textbf{Model} &
      \multicolumn{4}{c}{\textbf{WorldSense}} &
      \multicolumn{4}{c}{\textbf{VideoMMMU}} &
      \textbf{LRR} &
      \textbf{TVGBench} &
      \textbf{Average} \\
      \cmidrule(lr){2-5} \cmidrule(lr){6-9} \cmidrule(lr){10-10} \cmidrule(lr){11-11}
      &
      \textbf{Overall} &
      Recognition &
      Understanding &
      Reasoning &
      \textbf{Overall} &
      Comprehension &
      Adaptation &
      Perception &
      \textbf{Acc} &
      \textbf{mIoU} &
      \textbf{Score} \\
      \midrule
      \rowcolor{green!8} \multicolumn{12}{l}{\emph{Ablation Variants}} \\
      VISD w/o Feedback & 38.8 & 37.8 & 39.1 & 40.2 & 52.6 & 49.0 & 38.7 & 70.0 & 73.4 & 22.3 & 46.8 \\
      VISD w/ current-policy teacher & 39.2 & 37.8 & \textbf{40.6} & 40.1 & 52.0 & 48.7 & 39.0 & 68.3 & \textbf{75.3} & 18.7 & 46.3 \\
      VISD w/ Sync-10 teacher & 38.8 & 37.8 & 39.5 & 39.8 & 48.9 & 43.3 & 40.0 & 63.3 & 64.7 & 24.0 & 44.1 \\
      \textbf{VISD w/ EMA teacher (Ours)} & \textbf{39.9} & \textbf{39.4} & 40.4 & \textbf{40.5} & \textbf{53.8} & \textbf{49.7} & \textbf{42.3} & 69.3 & 73.5 & \textbf{24.8} & \textbf{48.0} \\
      \midrule
      \rowcolor{green!8} \multicolumn{12}{l}{\emph{Top-$K$ Analysis}} \\
      VISD w/ sampled token & 38.8 & 38.2 & 39.1 & 39.5 & \textbf{55.0} & \textbf{51.3} & \textbf{42.7} & \textbf{71.0} & 72.9 & 23.9 & 47.7 \\
      \textbf{VISD w/ top-$K$ (Ours)} & \textbf{39.9} & \textbf{39.4} & \textbf{40.4} & \textbf{40.5} & 53.8 & 49.7 & 42.3 & 69.3 & \textbf{73.5} & \textbf{24.8} & \textbf{48.0} \\
      \bottomrule[0.15em]
    \end{tabular}
  }
  \vspace{-4mm}
\end{table}

\paragraph{\normalfont\textbf{Structured feedback.}}
The no-feedback variant keeps the same policy-gradient objective and teacher replay, but conditions the teacher only on verified answer-side information.
Structured feedback improves final reward from 2.280 to 2.318 and answer accuracy from 0.565 to 0.571 (Figure~\ref{fig:training_ablation_curves}\textbf{(a)} and Table~\ref{tab:training_ablation}).
The same trend appears in Tables~\ref{tab:vstar_ablation} and~\ref{tab:general_ablation}, where the full VISD configuration obtains the best overall VSTAR score and the best average general-video score.
This suggests that the benefit is not merely from exposing the verified answer to the teacher; the judge feedback contributes diagnostic context that changes how token-level update magnitudes are distributed along the rollout.
Because the reinforcement advantage still determines the update direction, this extra context refines credit assignment without replacing the reward signal.
This mirrors the distinction made in recent self-distillation work between using richer feedback and performing denser token-level credit assignment: in VISD, the judge supplies the richer feedback, while teacher replay turns it into token-wise magnitude modulation.
Additional benchmark, training-curve, and mechanism analyses are provided in Appendices~\ref{app:benchmark_level_ablation_results}, \ref{app:additional_training_curves}, and~\ref{app:how_judge_credit}.

\paragraph{\normalfont\textbf{Teacher update strategy.}}
With judge feedback fixed, we compare a current-policy teacher, a Sync-10 teacher, and an EMA teacher (Figure~\ref{fig:training_ablation_curves}\textbf{(b)}).
The three variants are close while teacher guidance is active, suggesting that the early gain comes mainly from feedback-conditioned token credit rather than a particular synchronization trick.
After the teacher weight is annealed to zero, Sync-10 shows a sharp transient drop and a much larger final gradient norm, while EMA gives the best reward, answer accuracy, VSTAR overall score, and general-video average (Tables~\ref{tab:training_ablation}, \ref{tab:vstar_ablation}, and~\ref{tab:general_ablation}).
We therefore view EMA as a stabilizing parameterization of the feedback-conditioned teacher, not as the primary source of VISD's gains.
Additional optimization diagnostics are provided in Appendix~\ref{app:teacher_stability}.

\paragraph{\normalfont\textbf{Top-$K$ support vs. sampled token.}}
We compare VISD's teacher top-$K$ local support with a sampled-token variant that computes the teacher--student comparison only on the realized token.
The sampled-token variant is viable and remains close on several general video benchmarks, but top-$K$ maintains a stronger reward trajectory and better grounding scores, including VSTAR mAM/mLGM and TVGBench (Figure~\ref{fig:training_ablation_curves}\textbf{(c)}; Tables~\ref{tab:vstar_ablation} and~\ref{tab:general_ablation}).
This pattern is consistent with the motivation of local-support distillation: the teacher top-$K$ set gives a compact distribution-level comparison around the prefix, whereas the sampled-token variant observes only one realized point from that support.

\section{Conclusion}

In this work, we presented VISD, a structured self-distillation framework that addresses the challenge of fine-grained credit assignment in long-horizon video reasoning. Recognizing that errors in video QA and grounded reasoning are inherently structured—arising from temporal misalignment, logical inconsistency, or spatial grounding failures—VISD leverages a video-aware judge to generate diagnostically meaningful, token-level supervision. We integrate this privileged information into reinforcement learning through direction-magnitude decoupling mechanism, which preserves policy stability while enabling semantically aligned credit assignment. Supported by curriculum scheduling and EMA-based teacher stabilization, VISD robustly scales to complex video sequences. Extensive experiments demonstrate that VISD significantly accelerates convergence and improves accuracy, grounding, and interpretability over strong baselines. Future work will explore more efficient judge modeling and richer feedback structures for broader grounded reasoning applications.

\bibliographystyle{plainnat}
\bibliography{reference}


\clearpage
\appendix

\section*{Appendix Contents}
\startcontents[appendix]
\hypersetup{linkcolor=headingblue}
\printcontents[appendix]{l}{1}{\setcounter{tocdepth}{2}}
\hypersetup{linkcolor=seedblue}

\clearpage
\section{Extended Related Work}
\label{app:extended_related_work}

\paragraph{\normalfont\textbf{Video large language models.}}
Recent advances in video large language models (VideoLLMs) have extended the success of vision-language models to dynamic visual understanding~\cite{maaz2024videochatgpt,zhang2023videollama,zhang2025llavavideo,damonlpsg2025videollama3,liu2024st}.
Existing approaches typically build upon pretrained image-language backbones and incorporate temporal modeling through frame sampling~\cite{zhang2025llavavideo,damonlpsg2025videollama3}, temporal attention modules~\cite{wang2025internvideo2}, or lightweight adapters that aggregate information across frames~\cite{chen2024longvila,videoxl}.
These designs have enabled strong performance on a wide range of video tasks~\cite{fu2026videommev2stagebenchmarkscomprehensive,li2024mvbench,videommmu,cheng2025vstar,sun2024ppllava}, including question answering, captioning, and temporal reasoning.
Despite these improvements, most VideoLLMs are trained using supervised fine-tuning or imitation learning on static annotations~\cite{zhang2025llavavideo,damonlpsg2025videollama3}.
While effective for aligning model outputs with ground-truth answers, such training paradigms treat reasoning as a one-shot prediction problem and do not explicitly model the process of multi-step reasoning over time.
As a result, models often produce superficially correct answers that are not faithfully grounded in the underlying video content, especially in long or complex scenarios where errors may arise from incorrect temporal alignment or missing visual evidence~\cite{timer1,tvgr1,ouyang2025spacer,wang2025traceable}.
Several recent works attempt to address these limitations by incorporating iterative reasoning or external feedback signals~\cite{videochat-r1,wang2025videorft,videorts,thinkingwithvideos}.
However, these methods typically rely on heuristic supervision or unstructured feedback, which limits their ability to provide precise guidance for correcting intermediate reasoning steps.
Consequently, enabling fine-grained and interpretable credit assignment remains an open challenge in VideoLLM training.

\paragraph{\normalfont\textbf{Reinforcement learning for multimodal reasoning.}}
Reinforcement learning has emerged as a powerful paradigm for improving reasoning capabilities in large language models by optimizing policies against task-level rewards~\cite{deepseekai2026deepseekr1,grpo}.
In particular, reinforcement learning with verifiable rewards (RLVR) provides a principled way to incorporate correctness signals derived from ground-truth answers or external verifiers.
This paradigm has been increasingly applied to multimodal settings~\cite{feng2025videor1,videochat-r1,wang2025videorft}, where reward functions may jointly consider answer accuracy, grounding consistency, and other task-specific criteria.
While reinforcement learning offers reliable optimization signals, its effectiveness is fundamentally constrained by the granularity of reward design~\cite{deepseekai2026deepseekr1,grpo,yang2026selfdistilledrlvr}.
In most cases, rewards are defined at the sequence level, providing only a scalar signal for an entire reasoning trajectory.
This leads to coarse credit assignment, making it difficult to identify which parts of a long reasoning sequence are responsible for success or failure.
The problem is particularly severe in video reasoning, where errors are often localized and structured, such as referencing an incorrect time window or misidentifying objects in specific frames.
Recent efforts have explored combining reinforcement learning with auxiliary supervision to improve credit assignment~\cite{zhao2026selfdistilledreasoner,hubotter2026reinforcementlearning,yang2026selfdistilledrlvr}, including approaches that incorporate token-level guidance or policy regularization.
However, these methods typically treat auxiliary signals as generic constraints or smoothness priors, without explicitly modeling the structure of reasoning errors~\cite{zhao2026selfdistilledreasoner,hubotter2026reinforcementlearning,yang2026selfdistilledrlvr}.
As a result, they struggle to provide targeted corrections in complex multimodal scenarios.

\paragraph{\normalfont\textbf{Self-distillation and fine-grained credit assignment.}}
Self-distillation provides an alternative approach to improving sequence models by introducing dense, token-level supervision derived from teacher signals~\cite{hinton2015distill,bornagain,policy_distill}.
Among them, RLSD explores direction--magnitude decoupled self-distillation in scientific reasoning, while VISD adapts this principle to video reasoning, where feedback must account for answer correctness, reasoning consistency, temporal alignment, and spatial grounding.In the context of reasoning tasks, recent methods leverage self-distillation to refine intermediate predictions, often by comparing model outputs under different conditions or by integrating teacher policies into reinforcement learning frameworks~\cite{zhao2026selfdistilledreasoner,hubotter2026reinforcementlearning,yang2026selfdistilledrlvr}.
These approaches enable more fine-grained credit assignment than pure reinforcement learning, and have been shown to improve training efficiency and reasoning quality.
However, existing self-distillation methods largely rely on unstructured or modality-agnostic signals~\cite{zhao2026selfdistilledreasoner,hubotter2026reinforcementlearning,yang2026selfdistilledrlvr}.
The teacher guidance is typically expressed as a distributional difference or preference signal, without explicitly capturing the underlying causes of reasoning errors.
Unlike on-policy distillation methods that introduce dense distribution-matching or reverse-KL-style objectives, VISD does not add a separate distillation loss; the teacher signal only reweights the magnitude of the on-policy gradient while the reward advantage determines its direction.
This limitation becomes more pronounced in video reasoning, where failures may stem from diverse sources~\cite{meng2026openo3video,timer1,ouyang2025spacer,wang2025traceable}, including logical inconsistency, temporal misalignment, and incorrect visual grounding.
Without distinguishing between these factors, token-level supervision may lack diagnostic specificity, leading to suboptimal learning dynamics.
Furthermore, integrating self-distillation with reinforcement learning introduces additional challenges.
Naively combining the two can result in unstable optimization, as auxiliary signals may conflict with reward-driven updates or dominate the learning process.
This highlights the need for a principled framework that not only provides fine-grained supervision, but also aligns it with the underlying structure of multimodal reasoning and preserves the stability of policy optimization.

\section{Limitations and Future Works}
\label{app:limitations}

Our current evaluation mainly focuses on publicly available benchmarks for video reasoning and grounding.
While these benchmarks cover short-video reasoning, long-video understanding, temporal grounding, and spatial grounding, they still cannot fully capture the diversity and noise of real-world video use cases.
Future work can therefore extend VISD to more open-ended, noisy, and complex real-world video tasks, including settings with weaker supervision, more diverse video quality, and less standardized answer formats.

Our current method mainly focuses on video reasoning with explicit temporal and spatial grounding signals.
This setting is a natural testbed for studying fine-grained credit assignment, but it does not exhaust the full range of structured outputs that VideoLLMs may need to produce.
Future work can extend VISD to richer structured outputs, such as denser event descriptions or more compositional grounded reasoning formats, as well as more open-ended interactive video tasks.

\section{Implementation and Evaluation Details}
\label{app:implementation_details}

\subsection{Algorithm Details}
\label{app:algorithm_details}

\paragraph{\normalfont\textbf{Teacher and judge roles.}}
The feedback-conditioned teacher is initialized from the same SFT-initialized model as the student.
For each sampled student rollout, an external LLM judge reads the question, the student completion, the verified answer, and available grounding evidence, then produces structured process feedback about answer correctness, reasoning consistency, temporal alignment, and spatial grounding.
The judge uses these video-specific cues to diagnose the student's rollout, but the final training reward is still computed by the original verifier from the ground-truth answer and grounding annotations.
The full judge prompt template is provided in Appendix~\ref{app:judge_prompt}.
In the main experiments, we use GPT-5.4 as the judge because it provides strong instruction following and reasoning capabilities for diagnosing long video reasoning trajectories.
The judge is used only during training: it does not replace the verifier reward, does not provide gradients, and is not used at inference time.
Instead, its feedback is treated as privileged diagnostic context that tells the teacher not only whether a rollout is correct, but also how its reasoning or grounding succeeds or fails.
The teacher does not generate a new answer; it replays the same student completion under the hidden answer/evidence/feedback context and provides stop-gradient token scores.
The resulting teacher-student token score is used only to reweight the on-policy gradient magnitude, while the rollout-level advantage determines the update direction.

\paragraph{\normalfont\textbf{Reward.}}
The verifier reward is consistent with Open-o3-Video~\citep{meng2026openo3video} and VisionCoach~\citep{lee2026visioncoach}, covering answer correctness, format correctness, final-answer grounding, and reasoning-trace grounding.
VISD does not introduce a new reward source; instead, it changes how the resulting rollout reward is converted into token-level credit through feedback-conditioned teacher reweighting.
Each sampled rollout is scored by this verifier before computing the group-relative advantage.
For a prompt $x$, completion $y$, verified answer $a^\star$, and available grounding evidence $e$, the scalar reward is the sum of the applicable components:
\begin{equation}
  \begin{aligned}
  R(x,y,a^\star,e)
  &= \sum_{k\in\mathcal K(x)} \alpha_k r_k(x,y,a^\star,e),\\
  \mathcal K(x)
  &\subseteq
  \{\mathrm{ans},\mathrm{fmt},\mathrm{ans\text{-}tmp},
  \mathrm{ans\text{-}spa},\mathrm{thk\text{-}tmp},\mathrm{thk\text{-}spa}\}.
  \end{aligned}
  \label{eq:appendix_rollout_reward}
\end{equation}
Here $\mathcal K(x)$ masks out reward terms that are not supported by the current sample, and the main experiments use the configured component weights $\alpha_k$ when summing the rollout reward.
The answer reward checks exact correctness for multiple-choice questions and ROUGE-based similarity for free-form answers.
For temporal and spatial grounding answers, the verifier additionally computes temporal IoU between parsed answer intervals and ground-truth intervals, and visual IoU between parsed answer boxes and ground-truth boxes.
The format reward checks the structured response format, including paired \texttt{<think>} and \texttt{<answer>} tags and valid grounding tags such as \texttt{<obj>}, \texttt{<box>}, and \texttt{<t>} when grounding is required.

For reasoning traces, VISD parses timestamps and object-box-time tuples from the \texttt{<think>} section.
When interval supervision is available, the thinking temporal reward measures the fraction of parsed timestamps that fall inside the annotated interval:
\begin{equation}
  r_{\mathrm{thk\text{-}tmp}}^{\mathrm{seg}}
  =
  \frac{1}{M}\sum_{m=1}^{M}
  \mathbf{1}\!\left[s^\star \le t_m \le e^\star\right].
  \label{eq:appendix_temporal_segment_reward}
\end{equation}
When key-frame supervision is available, it instead uses an adaptive proximity score to the nearest annotated time:
\begin{equation}
  r_{\mathrm{thk\text{-}tmp}}^{\mathrm{pt}}
  =
  \frac{1}{M}\sum_{m=1}^{M}
  \exp\!\left(-\frac{\min_j |t_m-t_j^\star|^2}{2\sigma_s^2}\right),
  \label{eq:appendix_temporal_point_reward}
\end{equation}
where $\sigma_s$ is annealed during training so that early learning tolerates coarse temporal localization and later learning encourages tighter alignment.
For thinking spatial reward, a predicted tuple contributes only when its timestamp matches an annotated key frame within a small temporal tolerance; the reward then averages IoU scores between predicted boxes and ground-truth boxes on the matched frame:
\begin{equation}
  r_{\mathrm{thk\text{-}spa}}
  =
  \frac{1}{Z}\sum_{m\in\mathcal I}
  \max_{b^\star\in\mathcal B^\star(t_m)}
  \mathrm{IoU}(b_m,b^\star),
  \qquad
  \mathcal I=\{m:\min_j |t_m-t_j^\star|\le \tau\}.
  \label{eq:appendix_spatial_reward}
\end{equation}
If no valid prediction or supervision is available for a component, that component contributes zero or is masked out.
The summed rollout reward in Eq.~\eqref{eq:appendix_rollout_reward} is then used to compute the standard group-relative advantage in Eq.~\eqref{eq:appendix_grpo_advantage}; in the main VISD experiments, judge feedback and teacher replay do not create separate reward components or change the reward-determined advantage direction.

\paragraph{\normalfont\textbf{GRPO-style VISD objective.}}
For each prompt $x$, the old policy samples a group of rollouts $\{y^{(i)}\}_{i=1}^{G}$.
After scoring each rollout with the verifier, VISD first computes the standard group-relative advantage:
\begin{equation}
  A^{(i)} =
  \frac{R(x,y^{(i)})-\mu_G}{\sigma_G+\epsilon},
  \qquad
  \mu_G=\frac{1}{G}\sum_{j=1}^{G}R(x,y^{(j)}),
  \label{eq:appendix_grpo_advantage}
\end{equation}
where $\sigma_G$ is the standard deviation of rewards within the same rollout group.
At token position $t$ of rollout $i$, let $\Delta^{(i)}_t$ denote the stop-gradient top-$K$ teacher-student log-ratio on the realized token, as defined in Appendix~\ref{app:topk_local_support_ratio}.
VISD converts this ratio into a positive token modifier and reweighted advantage:
\begin{equation}
  m^{(i)}_t =
  \operatorname{clip}\!\left(
  \exp\left(\operatorname{sign}(A^{(i)})\Delta^{(i)}_t\right),
  1-\epsilon_w,\,1+\epsilon_w\right),
  \qquad
  \hat A^{(i)}_t =
  A^{(i)}\left[(1-\lambda_s)+\lambda_s m^{(i)}_t\right].
  \label{eq:appendix_visd_token_advantage}
\end{equation}
The final policy objective keeps the GRPO clipped surrogate form, but replaces the rollout-level advantage with the token-reweighted advantage:
\begin{equation}
  \begin{aligned}
  \mathcal J_{\mathrm{VISD}}(\theta)
  =
  \mathbb E\bigg[
  \frac{1}{G}\sum_{i=1}^{G}
  \frac{1}{|y^{(i)}|}\sum_{t=1}^{|y^{(i)}|}
  \min\Big(
  &\rho^{(i)}_t(\theta)\hat A^{(i)}_t,\\
  &\operatorname{clip}\!\left(\rho^{(i)}_t(\theta),1-\epsilon_{\mathrm{pg}},1+\epsilon_{\mathrm{pg}}\right)
  \hat A^{(i)}_t
  \Big)
  \bigg].
  \end{aligned}
  \label{eq:appendix_visd_objective}
\end{equation}
Here the token-level importance ratio is
\begin{equation}
  \rho^{(i)}_t(\theta) =
  \frac{\pi_\theta(y^{(i)}_t\mid x,y^{(i)}_{<t})}
       {\pi_{\theta_{\mathrm{old}}}(y^{(i)}_t\mid x,y^{(i)}_{<t})}.
  \label{eq:appendix_visd_ratio}
\end{equation}
The implementation minimizes $-\mathcal J_{\mathrm{VISD}}$.
When $\lambda_s=0$, $\hat A^{(i)}_t=A^{(i)}$, so VISD recovers the standard GRPO clipped surrogate.

\begin{algorithm}[h]
\small
\caption{VISD training procedure with feedback-conditioned teacher replay}
\label{alg:visd}
\begin{algorithmic}[1]
\Require Student policy $\pi_\theta$, EMA teacher $\pi_{\bar\theta}$, RL data $\mathcal{D}=\{(x,a^\star,e)\}$, judge $J$, verifier $R$, group size $G$, top-$K$ size $K$, VISD mixing schedule $\lambda_s$, credit clip bound $\epsilon_w$
\For{each training iteration $s$}
    \State Set rollout policy $\theta_{\mathrm{old}}\leftarrow\theta$
    \State Sample a batch of video-question prompts $\{x\}$ from $\mathcal{D}$
    \For{each prompt $x$ with verified answer/evidence $(a^\star,e)$}
        \Statex \textcolor{gray}{// Step 1: on-policy video reasoning rollouts}
        \State Sample $G$ completions $y^{(1)},\ldots,y^{(G)} \sim \pi_{\theta_{\mathrm{old}}}(\cdot\mid x)$
        \Statex \textcolor{gray}{// Step 2: rollout-level reward and update direction}
        \For{$i=1,\ldots,G$}
            \State Obtain reward $R^{(i)} \leftarrow R(x,y^{(i)},a^\star,e)$
        \EndFor
        \State Compute group-relative advantages $A^{(i)} \leftarrow (R^{(i)}-\mu_G)/(\sigma_G+\epsilon)$
        \Statex \textcolor{gray}{// Step 3: trajectory-dependent diagnosis}
        \For{$i=1,\ldots,G$}
            \State Generate structured feedback $f^{(i)} \leftarrow J(x,y^{(i)},a^\star,e)$
            \State Set privileged context $r^{(i)} \leftarrow (a^\star,e,f^{(i)})$ and replay the same completion $y^{(i)}$
            \For{$t=1,\ldots,|y^{(i)}|$}
                \State Construct $\mathcal{U}^{(i)}_t \leftarrow \operatorname{TopK}_{P_T(\cdot\mid x,r^{(i)},y^{(i)}_{<t})}(K)\cup\{y^{(i)}_t\}$
                \State Renormalize teacher and student distributions on $\mathcal{U}^{(i)}_t$
                \State Compute the top-$K$ support log-ratio as in Eq.~\eqref{eq: discrepancy}
                \Statex \hspace{\algorithmicindent}$\Delta^{(i)}_t \leftarrow \mathrm{sg}\!\left[\log \widetilde P_T(y^{(i)}_t\mid x,r^{(i)},y^{(i)}_{<t})-\log \widetilde P_S(y^{(i)}_t\mid x,y^{(i)}_{<t})\right]$
                \State Convert to clipped token weight $m^{(i)}_t \leftarrow \mathrm{clip}\!\left(\exp(\mathrm{sign}(A^{(i)})\Delta^{(i)}_t),1-\epsilon_w,1+\epsilon_w\right)$
                \State Set token advantage $\hat{A}^{(i)}_t \leftarrow A^{(i)}\!\cdot\!\left((1-\lambda_s)+\lambda_s m^{(i)}_t\right)$
            \EndFor
        \EndFor
    \EndFor
    \Statex \textcolor{gray}{// Step 4: direction-preserving policy update}
    \State Update $\theta$ with the clipped policy-gradient objective using $\{\hat{A}^{(i)}_t\}$
    \State Update EMA teacher $\bar\theta \leftarrow \tau\bar\theta+(1-\tau)\theta$ and anneal $\lambda_s$
\EndFor
\end{algorithmic}
\end{algorithm}

\subsection{Training Details}
\label{app:training_details}

\paragraph{\normalfont\textbf{Training data and initialization.}}
VISD uses the same SFT initialization recipe as Open-o3-Video, and our main method is applied in the subsequent RL stage.
The SFT initialization uses STGR-CoT-30k, which contains four groups of grounded reasoning data: temporal grounding CoT samples, spatial grounding samples, spatio-temporal reasoning samples curated from temporal grounding and dense-captioning sources, and general video reasoning CoT samples.
This initialization stage teaches the model to produce the grounded reasoning format used throughout our experiments, including temporal references, object mentions, bounding boxes, and final answers.
Starting from this SFT-initialized model, we further optimize VISD on STGR-RL-36k with on-policy rollouts.
The RL data follows the same broad task coverage and includes temporal grounding, spatial grounding, spatio-temporal reasoning, and general video QA prompts.
Each RL example contains a video, a question, a verified answer, and, when available, grounding-related metadata such as key frames, key objects, temporal segments, or spatial regions.
The RL stage optimizes answer correctness and spatio-temporal grounding under task-specific rewards, while VISD introduces judge feedback and teacher replay as training-time privileged signals for token-level credit reweighting.

\paragraph{\normalfont\textbf{Optimization details.}}
We train VISD on 2 nodes with 8 NVIDIA A100 GPUs per node, using DeepSpeed ZeRO-3 and FlashAttention-2.
Both the SFT initialization stage and the RL stage are trained for one epoch with learning rate $1\times10^{-6}$.
For each prompt, the policy samples $G=4$ on-policy completions.
The maximum prompt length is 16{,}384 tokens and the maximum completion length is 768 tokens.
We use a per-device training batch size of 1, gradient accumulation of 1, learning rate $1\times10^{-6}$, weight decay 0.01, and gradient clipping with maximum norm 5.
The maximum visual input resolution is controlled by a pixel budget of 401{,}408 pixels during training.
The main VISD result evaluates the student policy from the EMA-teacher training setting at optimization step 1200.

\paragraph{\normalfont\textbf{VISD training configuration.}}
While teacher guidance is active, we maintain an EMA teacher and update it every optimization step with update rate 0.01.
We do not use a reference model, a teacher KL loss, or a separate distillation loss in the main VISD setting.
The VISD top-$K$ support size is $K=16$, and the token weight is clipped to $[0.8,1.2]$.
The VISD mixing coefficient starts from 0.5 and linearly anneals to 0 over the first 600 optimization steps, gradually reducing the objective to standard on-policy RL.
Table~\ref{tab:appendix_training_hyperparams} summarizes the main model configuration and optimization hyperparameters used in VISD.

\begin{table}[h]
  \vspace{-2mm}
  \centering
  \small
  \caption{\small \emph{Main model configuration and training hyperparameters for VISD.}
  The table summarizes the backbone, teacher/evaluator setup, rollout configuration, token-credit reweighting, and optimization hyperparameters.}
  \label{tab:appendix_training_hyperparams}
  \begin{tabular}{lc}
    \toprule
    Parameter & Value \\
    \midrule
    \multicolumn{2}{l}{\textbf{Model, teacher, and evaluator}} \\
    Base VideoLLM & Qwen2.5-VL-7B \\
    Teacher initialization & Same as student initialization \\
    Teacher update & EMA, update rate 0.01 \\
    VSTAR answer evaluator & Qwen2.5-72B-Instruct \\
    \midrule
    \multicolumn{2}{l}{\textbf{Rollout and sequence length}} \\
    Rollouts per prompt & 4 \\
    Max prompt length & 16{,}384 \\
    Max completion length & 768 \\
    Training pixel budget & 401{,}408 \\
    \midrule
    \multicolumn{2}{l}{\textbf{VISD credit reweighting}} \\
    Top-$K$ support size & 16 \\
    Token-weight clip range & $[0.8,1.2]$ \\
    Initial VISD mixing coefficient & 0.5 \\
    Mixing annealing steps & 600 \\
    Reference-model KL & Not used \\
    Auxiliary distillation loss & Not used \\
    \midrule
    \multicolumn{2}{l}{\textbf{Optimization}} \\
    SFT epochs & 1 \\
    RL epochs & 1 \\
    Per-device batch size & 1 \\
    Gradient accumulation & 1 \\
    Optimizer & AdamW \\
    Learning rate & $1\times10^{-6}$ \\
    Weight decay & 0.01 \\
    Max gradient norm & 5 \\
    Precision & bfloat16 \\
    \bottomrule
  \end{tabular}
  \vspace{-2mm}
\end{table}

\subsection{Evaluation Details}
\label{app:evaluation_details}

\paragraph{\normalfont\textbf{Baselines.}}
We compare VISD with both proprietary and open-source video reasoning models.
For VSTAR, we include proprietary models such as GPT-4o~\citep{gpt4o} and Gemini-2-Flash~\citep{comanici2025gemini}, as well as open-source VideoLLMs including TRACE~\citep{guo2024trace}, Oryx-1.5-7B~\citep{liu2024oryx}, VideoChat2~\citep{li2024mvbench}, Qwen2.5-VL-7B~\citep{qwen2.5vl}, InternVL-2.5-8B~\citep{internvl2.5}, Video-LLaMA3~\citep{damonlpsg2025videollama3}, LLaVA-Video~\citep{zhang2025llavavideo}, Open-o3-Video~\citep{meng2026openo3video}, and VisionCoach~\citep{lee2026visioncoach}.
For broader video understanding benchmarks, we additionally compare against recent reasoning-oriented or tool-based baselines, including LongVT-RL~\citep{yang2025longvt}, VideoRFT~\citep{wang2025videorft}, and Video-R1~\citep{feng2025videor1}.
For zero-shot temporal grounding on Charades-STA~\citep{gao2017tall}, we report representative general VideoLLMs and temporal-grounding VideoLLMs, following the standard comparison protocol used in prior work.
Rows marked with $^*$ are evaluated by us under the same inference pipeline, while rows marked with $^\dagger$ combine available reported results with our own evaluations when necessary.

\paragraph{\normalfont\textbf{VSTAR metrics.}}
VSTAR evaluates grounded video reasoning along three dimensions: \textit{what}, \textit{when}, and \textit{where}.
The \textit{what} dimension is measured by answer accuracy.
The \textit{when} dimension is measured by temporal IoU between predicted and ground-truth temporal intervals.
The \textit{where} dimension is measured by spatial IoU between predicted and ground-truth bounding boxes.
VSTAR reports two reasoning chains: Chain1 follows the order \textit{what--when--where}, while Chain2 follows \textit{what--where--when}.
For each chain, the arithmetic mean is computed as
\begin{equation}
  \mathrm{AM} = \frac{1}{3}\left(\mathrm{Acc}+\mathrm{tIoU}+\mathrm{sIoU}\right),
  \label{eq:appendix_vstar_am}
\end{equation}
and VSTAR further reports mAM by averaging AM over the two chains.
The benchmark also reports mLGM, a logarithmic geometric-style aggregate that reduces the collapse caused by any single near-zero grounding dimension.
Together, mAM and mLGM summarize answer correctness, temporal localization, and spatial grounding into overall spatio-temporal reasoning scores.

\paragraph{\normalfont\textbf{General video understanding metrics.}}
For WorldSense~\citep{worldsense}, VideoMMMU~\citep{videommmu}, Video-MME-v2~\citep{fu2026videommev2stagebenchmarkscomprehensive}, and LongVideo-Reason~\citep{longvila-r1}, we follow the official benchmark protocols and report the benchmark-level accuracy or score used by each dataset.
For WorldSense and VideoMMMU, we additionally report their category-level scores when available.
For Video-MME-v2~\citep{fu2026videommev2stagebenchmarkscomprehensive}, we report the official non-linear score, level-wise scores, consistency, coherence, and average accuracy using 64 sampled frames without subtitles.

\paragraph{\normalfont\textbf{Temporal grounding metrics.}}
For TVGBench~\citep{timer1} and Charades-STA~\citep{gao2017tall}, we evaluate temporal grounding using temporal Intersection over Union (tIoU) between the predicted and ground-truth intervals.
For Charades-STA, we report Recall at IoU thresholds 0.3, 0.5, and 0.7, together with mean IoU.
For TVGBench, we report mean IoU following the benchmark protocol.

\paragraph{\normalfont\textbf{Evaluation protocol.}}
All evaluations are conducted on 8 NVIDIA A100 GPUs.
For VSTAR~\citep{cheng2025vstar}, we follow the official evaluation pipeline with 16 sampled frames and report answer accuracy, temporal IoU, spatial IoU, mAM, and mLGM.
The generated VSTAR answers are evaluated by a Qwen2.5-72B-Instruct judge following the benchmark protocol.
For Video-MME-v2~\citep{fu2026videommev2stagebenchmarkscomprehensive}, we sample 64 frames without subtitles and report the official non-linear score, level-wise scores, consistency, coherence, and average accuracy.
For WorldSense~\citep{worldsense}, VideoMMMU~\citep{videommmu}, Video-MME-v2~\citep{fu2026videommev2stagebenchmarkscomprehensive}, LongVideo-Reason~\citep{longvila-r1}, TVGBench~\citep{timer1}, and Charades-STA~\citep{gao2017tall}, we use the corresponding official or benchmark-provided evaluation scripts, keeping decoding deterministic for all evaluated models.

\section{Additional Experimental Results}
\label{app:benchmark_ablation_results}

\subsection{Benchmark-Level Ablation Results}
\label{app:benchmark_level_ablation_results}

Tables~\ref{tab:appendix_videommev2_ablation} and~\ref{tab:appendix_charades_sta_ablation} provide additional benchmark-level ablation results for Video-MME-v2, Charades-STA, and TVGBench.
The VSTAR and general video reasoning ablation results are reported in Section~\ref{sec:ablation}.
These results complement the training curves in Section~\ref{sec:ablation}: the benchmark scores are not meant to replace the main comparison, but to check whether the same design choices remain visible under downstream evaluation.

\begin{table}[h]
  \vspace{-2mm}
  \centering
  \small
  \caption{\small \emph{Video-MME-v2 ablation results.}
  All evaluated ablation models use 64 sampled frames without subtitles. We separate teacher/feedback ablations from the top-$K$ analysis because they vary different design axes.}
  \label{tab:appendix_videommev2_ablation}
  \resizebox{\linewidth}{!}{
    \begin{tabular}{lcccccccc}
      \toprule[0.15em]
      \textbf{Model} & \textbf{Frames} & \textbf{Non-Lin Score} & \textbf{Level 1} & \textbf{Level 2} & \textbf{Level 3} & \textbf{Consistency} & \textbf{Coherence} & \textbf{Avg Acc} \\
      \midrule
      \rowcolor{green!8} \multicolumn{9}{l}{\emph{Ablation Variants}} \\
      VISD w/o Feedback & 64 & 7.6 & 11.0 & 6.4 & 6.5 & 8.1 & 6.7 & 20.9 \\
      VISD w/ current-policy teacher & 64 & 7.7 & 10.2 & 7.0 & 6.7 & 8.0 & 6.9 & 21.2 \\
      VISD w/ Sync-10 teacher & 64 & 8.1 & 12.6 & 6.7 & 6.5 & 9.1 & 6.3 & 21.7 \\
      \textbf{VISD w/ EMA teacher (Ours)} & 64 & \textbf{9.4} & \textbf{13.4} & \textbf{8.2} & \textbf{7.9} & \textbf{10.2} & \textbf{8.0} & \textbf{23.8} \\
      \midrule
      \rowcolor{green!8} \multicolumn{9}{l}{\emph{Top-$K$ Analysis}} \\
      VISD w/ sampled token & 64 & 8.1 & 11.9 & 6.7 & 6.9 & 8.6 & 7.2 & 21.5 \\
      \textbf{VISD w/ top-$K$ (Ours)} & 64 & \textbf{9.4} & \textbf{13.4} & \textbf{8.2} & \textbf{7.9} & \textbf{10.2} & \textbf{8.0} & \textbf{23.8} \\
      \bottomrule[0.15em]
    \end{tabular}
  }
\end{table}

\begin{table}[h]
  \vspace{-2mm}
  \centering
  \small
  \caption{\small \emph{Detailed temporal-grounding ablation results.}
  We report Charades-STA and TVGBench temporal-grounding metrics. Teacher/feedback ablations and top-$K$ analysis are shown as separate blocks.}
  \label{tab:appendix_charades_sta_ablation}
  \resizebox{\linewidth}{!}{
    \begin{tabular}{lcccccccc}
      \toprule[0.15em]
      \textbf{Model} &
      \multicolumn{4}{c}{\textbf{Charades-STA}} &
      \multicolumn{4}{c}{\textbf{TVGBench}} \\
      \cmidrule(lr){2-5} \cmidrule(lr){6-9}
      &
      \textbf{R@0.3} &
      \textbf{R@0.5} &
      \textbf{R@0.7} &
      \textbf{mIoU} &
      \textbf{R@0.3} &
      \textbf{R@0.5} &
      \textbf{R@0.7} &
      \textbf{mIoU} \\
      \midrule
      \rowcolor{green!8} \multicolumn{9}{l}{\emph{Ablation Variants}} \\
      VISD w/o Feedback & 60.5 & 40.7 & 18.5 & 40.2 & 32.5 & 21.4 & 11.8 & 22.3 \\
      VISD w/ current-policy teacher & 48.8 & 27.5 & 11.2 & 31.8 & 28.2 & 16.4 & 7.0 & 18.7 \\
      VISD w/ Sync-10 teacher & 66.6 & 42.8 & 19.2 & 43.2 & \textbf{35.6} & 22.2 & 10.4 & 24.0 \\
      \textbf{VISD w/ EMA teacher (Ours)} & \textbf{67.3} & \textbf{46.1} & \textbf{21.6} & \textbf{44.6} & 34.5 & \textbf{24.5} & \textbf{13.4} & \textbf{24.8} \\
      \midrule
      \rowcolor{green!8} \multicolumn{9}{l}{\emph{Top-$K$ Analysis}} \\
      VISD w/ sampled token & 63.6 & 41.4 & 18.8 & 41.7 & 35.0 & 21.5 & 12.8 & 23.9 \\
      \textbf{VISD w/ top-$K$ (Ours)} & \textbf{67.3} & \textbf{46.1} & \textbf{21.6} & \textbf{44.6} & 34.5 & \textbf{24.5} & \textbf{13.4} & \textbf{24.8} \\
      \bottomrule[0.15em]
    \end{tabular}
  }
  \vspace{-2mm}
\end{table}

\subsection{Additional Training Curves}
\label{app:additional_training_curves}

\begin{figure}[htbp]
  \vspace{-2mm}
  \centering
  \begin{minipage}{0.32\linewidth}
    \centering
    \includegraphics[width=\linewidth]{images/training_curve/fig_ablation_topk_sampled_total_reward.pdf}\\[-0.2em]
    {\footnotesize\textbf{(a)}}
  \end{minipage}\hfill
  \begin{minipage}{0.32\linewidth}
    \centering
    \includegraphics[width=\linewidth]{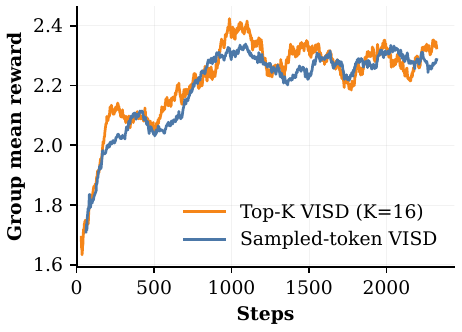}\\[-0.2em]
    {\footnotesize\textbf{(b)}}
  \end{minipage}\hfill
  \begin{minipage}{0.32\linewidth}
    \centering
    \includegraphics[width=\linewidth]{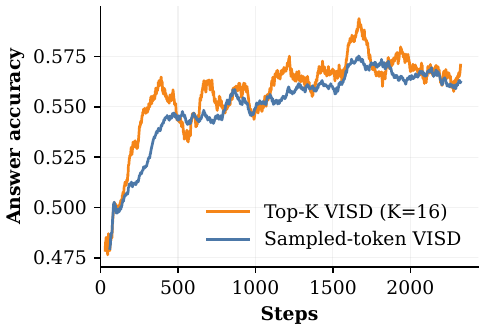}\\[-0.2em]
    {\footnotesize\textbf{(c)}}
  \end{minipage}

  \vspace{0.4em}
  \begin{minipage}{0.32\linewidth}
    \centering
    \includegraphics[width=\linewidth]{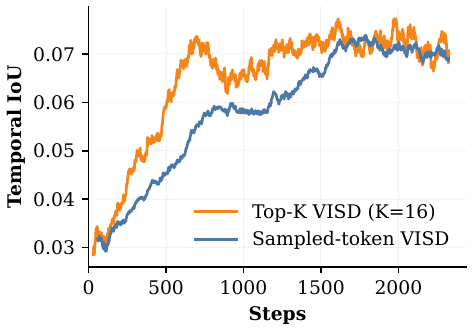}\\[-0.2em]
    {\footnotesize\textbf{(d)}}
  \end{minipage}\hfill
  \begin{minipage}{0.32\linewidth}
    \centering
    \includegraphics[width=\linewidth]{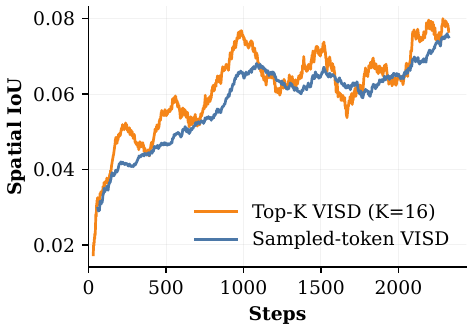}\\[-0.2em]
    {\footnotesize\textbf{(e)}}
  \end{minipage}\hfill
  \begin{minipage}{0.32\linewidth}
    \centering
    \includegraphics[width=\linewidth]{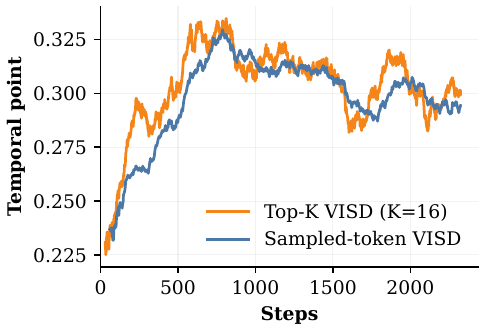}\\[-0.2em]
    {\footnotesize\textbf{(f)}}
  \end{minipage}

  \vspace{0.4em}
  \begin{minipage}{0.32\linewidth}
    \centering
    \includegraphics[width=\linewidth]{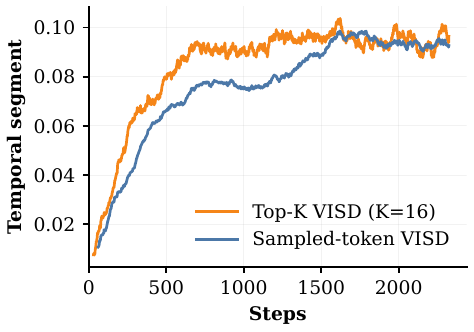}\\[-0.2em]
    {\footnotesize\textbf{(g)}}
  \end{minipage}\hfill
  \begin{minipage}{0.32\linewidth}
    \centering
    \includegraphics[width=\linewidth]{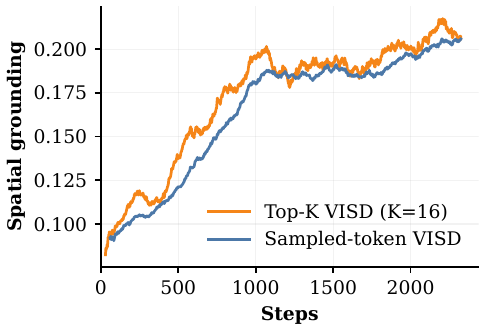}\\[-0.2em]
    {\footnotesize\textbf{(h)}}
  \end{minipage}\hfill
  \begin{minipage}{0.32\linewidth}
    \centering
    \includegraphics[width=\linewidth]{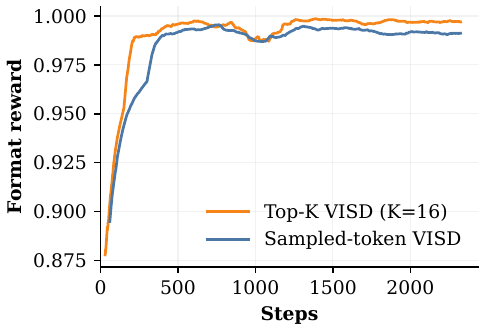}\\[-0.2em]
    {\footnotesize\textbf{(i)}}
  \end{minipage}
  \vspace{-2mm}
  \caption{\small \emph{Top-$K$ support versus sampled-token training curves.}
  We compare \textbf{(a)} total reward, \textbf{(b)} group mean reward, \textbf{(c)} answer accuracy,
  \textbf{(d)} temporal IoU, \textbf{(e)} spatial IoU, \textbf{(f)} temporal point,
  \textbf{(g)} temporal segment, \textbf{(h)} spatial grounding, and \textbf{(i)} format reward.}
  \label{fig:appendix_topk_sampled_curves}
  \vspace{-2mm}
\end{figure}

\paragraph{\normalfont\textbf{Top-$K$ support versus sampled token.}}
Figure~\ref{fig:appendix_topk_sampled_curves} provides the component-level training curves for the top-$K$ support and sampled-token variants discussed in Section~\ref{sec:ablation}.
The sampled-token variant remains competitive on several signals, but the top-$K$ support gives a slightly stronger and more stable trend on total reward and several grounding-related components, matching the benchmark-level differences in Tables~\ref{tab:vstar_ablation}, \ref{tab:general_ablation}, \ref{tab:appendix_videommev2_ablation}, and~\ref{tab:appendix_charades_sta_ablation}.

\begin{figure}[htbp]
  \vspace{-2mm}
  \centering
  \begin{minipage}{0.32\linewidth}
    \centering
    \includegraphics[width=\linewidth]{images/training_curve/fig_ablation_feedback_total_reward.pdf}\\[-0.2em]
    {\footnotesize\textbf{(a)}}
  \end{minipage}\hfill
  \begin{minipage}{0.32\linewidth}
    \centering
    \includegraphics[width=\linewidth]{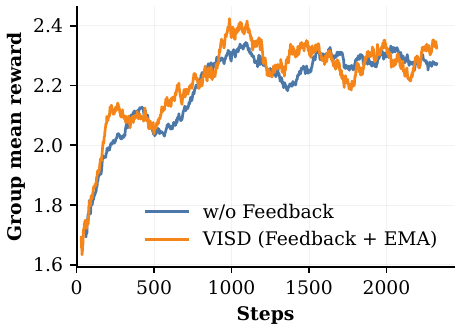}\\[-0.2em]
    {\footnotesize\textbf{(b)}}
  \end{minipage}\hfill
  \begin{minipage}{0.32\linewidth}
    \centering
    \includegraphics[width=\linewidth]{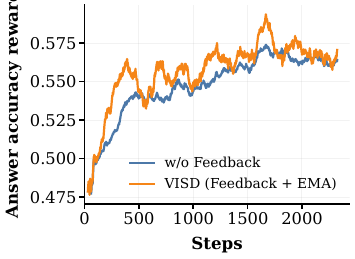}\\[-0.2em]
    {\footnotesize\textbf{(c)}}
  \end{minipage}

  \vspace{0.4em}
  \begin{minipage}{0.32\linewidth}
    \centering
    \includegraphics[width=\linewidth]{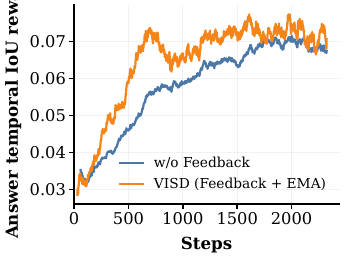}\\[-0.2em]
    {\footnotesize\textbf{(d)}}
  \end{minipage}\hfill
  \begin{minipage}{0.32\linewidth}
    \centering
    \includegraphics[width=\linewidth]{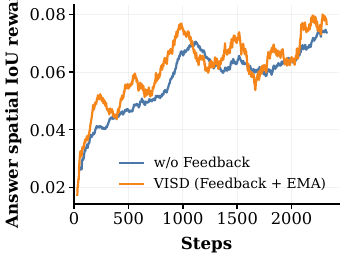}\\[-0.2em]
    {\footnotesize\textbf{(e)}}
  \end{minipage}\hfill
  \begin{minipage}{0.32\linewidth}
    \centering
    \includegraphics[width=\linewidth]{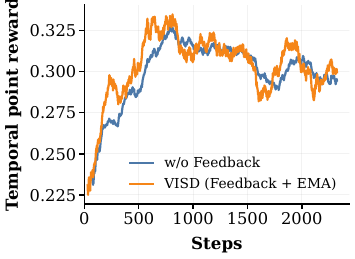}\\[-0.2em]
    {\footnotesize\textbf{(f)}}
  \end{minipage}

  \vspace{0.4em}
  \begin{minipage}{0.32\linewidth}
    \centering
    \includegraphics[width=\linewidth]{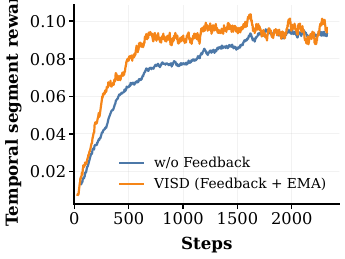}\\[-0.2em]
    {\footnotesize\textbf{(g)}}
  \end{minipage}\hfill
  \begin{minipage}{0.32\linewidth}
    \centering
    \includegraphics[width=\linewidth]{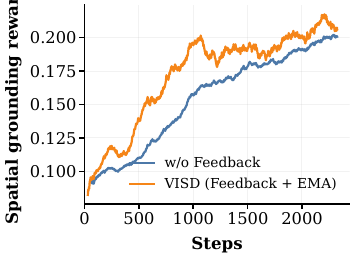}\\[-0.2em]
    {\footnotesize\textbf{(h)}}
  \end{minipage}\hfill
  \begin{minipage}{0.32\linewidth}
    \centering
    \includegraphics[width=\linewidth]{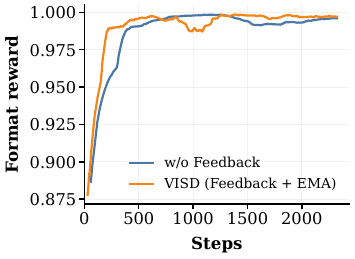}\\[-0.2em]
    {\footnotesize\textbf{(i)}}
  \end{minipage}
  \vspace{-2mm}
  \caption{\small \emph{Component-level reward curves for feedback ablation.}
  Each panel compares VISD with and without judge feedback for one reward component:
  \textbf{(a)} total reward, \textbf{(b)} group mean reward, \textbf{(c)} answer accuracy,
  \textbf{(d)} temporal IoU, \textbf{(e)} spatial IoU, \textbf{(f)} temporal point,
  \textbf{(g)} temporal segment, \textbf{(h)} spatial grounding, and \textbf{(i)} format.}
  \label{fig:appendix_reward_components_split}
  \vspace{-2mm}
\end{figure}

\paragraph{\normalfont\textbf{Feedback conditioning.}}
Figure~\ref{fig:appendix_reward_components_split} compares the feedback and no-feedback variants across individual reward components.
We report each component as a separate small panel rather than a dense single-axis multi-panel plot for readability.
These curves are noisier than the total reward curve, especially for sparse grounding sub-rewards, so we keep them in the appendix.
They are nevertheless useful as a sanity check: VISD's improvement is not caused by a single formatting artifact, but is accompanied by stronger answer and grounding-related signals over training.

\subsection{Teacher Stability Diagnostics}
\label{app:teacher_stability}

\begin{figure}[htbp]
  \centering
  \begin{minipage}{0.48\linewidth}
    \centering
    \includegraphics[width=\linewidth]{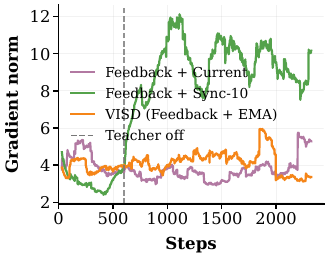}\\[-0.2em]
    {\footnotesize\textbf{(a)}}
  \end{minipage}\hfill
  \begin{minipage}{0.48\linewidth}
    \centering
    \includegraphics[width=\linewidth]{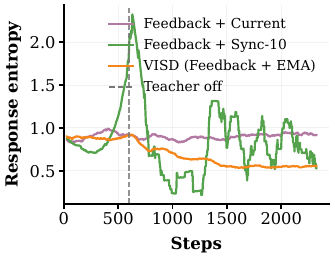}\\[-0.2em]
    {\footnotesize\textbf{(b)}}
  \end{minipage}
  \caption{\small \emph{Teacher update stability diagnostics.}
  We show \textbf{(a)} gradient norm and \textbf{(b)} response entropy for different teacher parameterizations.}
  \label{fig:appendix_teacher_stability_split}
\end{figure}

Figure~\ref{fig:appendix_teacher_stability_split} provides additional optimization diagnostics for the teacher update ablation.
The Sync-10 teacher synchronizes parameters with the student every 10 steps and keeps them frozen between synchronizations, whereas the current-policy teacher uses the instantaneous student parameters.
The EMA teacher smooths high-frequency parameter changes and gives a more stable teacher signal.
Consistent with Table~\ref{tab:training_ablation}, the Sync-10 variant exhibits a substantially larger gradient norm and a sharper disturbance as the teacher weight approaches zero, while EMA preserves the best final reward.

\subsection{Training Details for Ablation Settings}
\label{app:ablation_details}

The ablation curves are plotted by training step.
For the feedback comparison, the no-feedback setting uses the same reinforcement-learning and teacher-replay pipeline but removes the judge-generated process feedback.
Teacher-weight annotations are shown only in the teacher-update comparison, where all compared settings share the same teacher schedule.

For the teacher-update ablation, all compared variants use judge feedback.
The current-policy teacher directly uses the latest student parameters.
The Sync-10 teacher synchronizes teacher parameters with the student every 10 training steps and keeps them frozen between synchronizations.
The EMA teacher updates teacher parameters as an exponential moving average of the student parameters.
All reported final metrics in Table~\ref{tab:training_ablation} are averaged over the final 100 training steps.

\section{Mechanistic Analysis and Qualitative Diagnostics}
\label{app:mechanistic_analysis}

\subsection{How Judge Feedback Affects Credit Assignment}
\label{app:how_judge_credit}

The ablation in Section~\ref{sec:ablation} shows that judge feedback improves training, but it does not by itself explain \emph{how} the feedback enters the policy update.
VISD uses the judge as a trajectory-level diagnostic channel.
As illustrated in Appendix~\ref{app:judge_feedback_case}, different rollouts of the same question receive different judge evaluations: the judge distinguishes failed, partially correct, and correct trajectories, and points to whether the error comes from the answer, temporal coverage, spatial grounding, or the consistency between reasoning and answer.
This suggests that the feedback is not merely an additional prompt field, but a rollout-dependent description of \emph{how} the attempt succeeds or fails.

The teacher then converts this diagnosis into token-level magnitude modulation.
An answer-only teacher observes the verified answer-side information, while the feedback-conditioned teacher additionally observes the judge diagnosis.
Thus, the teacher replay distribution changes from $q(\cdot \mid x, a^\star, y_{<t})$ to $q(\cdot \mid x, a^\star, f, y_{<t})$, where $f$ denotes judge feedback.
The rollout-level advantage still determines whether the whole trajectory should be reinforced or suppressed; the teacher-student ratio only modulates the relative update magnitude of individual tokens.
This direction--magnitude separation prevents the judge from overwriting the reward direction, while still allowing feedback to redistribute learning pressure within the trajectory.

We provide a formal decomposition and token-level visualizations in Appendix~\ref{app:feedback_conditioned_credit}.
The qualitative comparison shows that feedback-conditioned replay does not uniformly change all tokens relative to answer-only replay.
Instead, the additional changes are sparse and tend to appear around tokens tied to diagnosed process errors, such as spurious actions, incomplete temporal coverage, or answer-critical object descriptions.
These visualizations are intended as mechanism evidence rather than standalone performance evidence: they illustrate how the extra information in judge feedback can influence credit assignment over process tokens on top of answer-only supervision.

\subsection{Trajectory-Dependent Judge Feedback}
\label{app:judge_feedback_case}

\begin{figure}[h]
  \centering
  \includegraphics[width=0.98\linewidth]{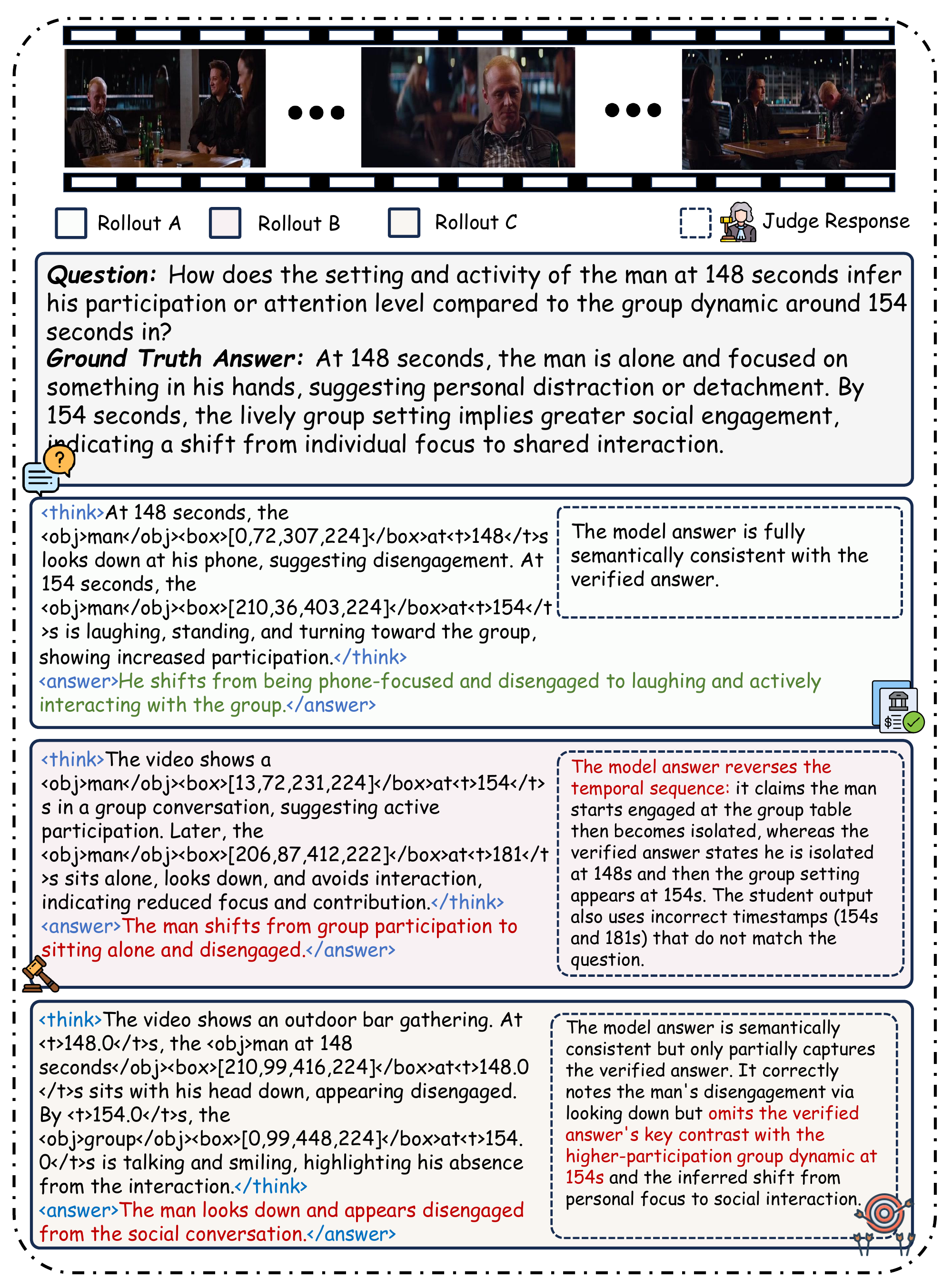}
  \caption{\small \emph{Trajectory-dependent judge feedback.}
  For the same video question, the judge provides different evaluations and diagnoses for different student rollouts.
  The feedback is used as privileged information for teacher replay rather than as a replacement for the reinforcement reward.}
  \label{fig:appendix_judge_feedback}
\end{figure}

Figure~\ref{fig:appendix_judge_feedback} shows the judge feedback for different rollouts of the same video question.
The judge assigns different evaluations to failed, partially correct, and correct trajectories, and provides diagnosis about the corresponding failure modes.
This example supports the role of the judge as a rollout-dependent diagnostic channel: the feedback describes \textbf{\emph{not only whether a rollout succeeds, but also how its reasoning, answer, and temporal evidence agree or conflict}}.
VISD uses this signal as privileged context for teacher replay, while the rollout-level reward advantage remains responsible for the sign of the policy update.

\subsection{Top-$K$ Local Support Ratio}
\label{app:topk_local_support_ratio}

This section clarifies the top-$K$ ratio used by VISD.
Let $q_t(v)$ denote the feedback-conditioned teacher distribution at token position $t$, and let $\pi_t(v)$ denote the student distribution under the ordinary rollout context:
\begin{equation}
  q_t(v) = q(v \mid x,a^\star,e,f,y_{<t}),
  \qquad
  \pi_t(v) = \pi_\theta(v \mid x,y_{<t}).
  \label{eq:appendix_teacher_student_dist}
\end{equation}
VISD constructs a teacher-centered local support by augmenting the teacher top-$K$ candidates with the realized rollout token:
\begin{equation}
  \mathcal U_t
  =
  \operatorname{TopK}_{q_t}(K)
  \cup
  \{y_t\}.
  \label{eq:appendix_topk_support}
\end{equation}
The augmentation by $y_t$ is not an additional sampled-token objective; it only ensures that the on-policy token receiving the policy-gradient update is always covered by the local support.
Within $\mathcal U_t$, teacher and student probabilities are renormalized as
\begin{equation}
  \widetilde q_t(v)
  =
  \frac{q_t(v)}
  {\sum_{u\in\mathcal U_t}q_t(u)},
  \qquad
  \widetilde \pi_t(v)
  =
  \frac{\pi_t(v)}
  {\sum_{u\in\mathcal U_t}\pi_t(u)},
  \qquad v\in\mathcal U_t .
  \label{eq:appendix_topk_renorm}
\end{equation}
VISD then computes the teacher-student evidence ratio on the realized token:
\begin{equation}
  \Delta_t^{K}
  =
  \operatorname{sg}
  \left[
  \log \widetilde q_t(y_t)
  -
  \log \widetilde \pi_t(y_t)
  \right].
  \label{eq:appendix_topk_delta}
\end{equation}
Unlike a top-$K$ KL loss, Eq.~\ref{eq:appendix_topk_delta} does not introduce gradients over all tokens in $\mathcal U_t$.
It only provides a stop-gradient scalar used to reweight the policy-gradient term of the generated token.

\paragraph{\normalfont\textbf{Local-support calibration.}}
Let
\begin{equation}
  Q_t(\mathcal U_t)=\sum_{u\in\mathcal U_t}q_t(u),
  \qquad
  P_t(\mathcal U_t)=\sum_{u\in\mathcal U_t}\pi_t(u).
  \label{eq:appendix_support_mass}
\end{equation}
Then Eq.~\ref{eq:appendix_topk_delta} admits the exact decomposition
\begin{equation}
  \Delta_t^{K}
  =
  \underbrace{
  \log q_t(y_t)-\log \pi_t(y_t)
  }_{\text{token evidence ratio}}
  +
  \underbrace{
  \log\frac{P_t(\mathcal U_t)}{Q_t(\mathcal U_t)}
  }_{\text{local-support calibration}} .
  \label{eq:appendix_topk_decomposition}
\end{equation}
This identity shows that the top-$K$ ratio preserves the original teacher-student evidence ratio on the realized token, while adding a calibration term that accounts for how much probability mass the teacher and student assign to the teacher-supported local region.

\paragraph{\normalfont\textbf{Direction preservation and bounded redistribution.}}
VISD converts the local-support ratio into a positive token weight:
\begin{equation}
  m_t
  =
  \operatorname{clip}
  \left(
  \exp\left(\operatorname{sign}(A)\Delta_t^{K}\right),
  1-\epsilon_w,
  1+\epsilon_w
  \right).
  \label{eq:appendix_topk_weight}
\end{equation}
The token-level advantage is
\begin{equation}
  \hat A_t
  =
  A\left[(1-\lambda)+\lambda m_t\right].
  \label{eq:appendix_topk_advantage}
\end{equation}
Because $m_t>0$, the reweighting cannot flip the rollout-level update direction:
\begin{equation}
  \operatorname{sign}(\hat A_t)=\operatorname{sign}(A),
  \qquad A\neq 0.
  \label{eq:appendix_topk_direction}
\end{equation}
Moreover, since $m_t\in[1-\epsilon_w,1+\epsilon_w]$,
\begin{equation}
  \frac{\hat A_t}{A}
  \in
  [1-\lambda\epsilon_w,\;1+\lambda\epsilon_w].
  \label{eq:appendix_bounded_redistribution}
\end{equation}
Thus, the teacher-conditioned signal only redistributes token-level update magnitudes within a bounded range, and the update naturally reduces to the standard rollout advantage as $\lambda$ anneals to zero.

\subsection{Feedback-Conditioned Credit as an Information Layer}
\label{app:feedback_conditioned_credit}

This section formalizes the difference between answer-only teacher replay and feedback-conditioned teacher replay.
Let $a^\star$ denote verified answer-side information and let $f$ denote judge feedback for a sampled student rollout.
At token position $t$, we compare the two teacher contexts on a common local support and let $\widetilde q_t^a$ and $\widetilde q_t^{a,f}$ denote the renormalized teacher distributions under answer-only and feedback-conditioned contexts, respectively.
The answer-only teacher-student log-ratio is
\begin{equation}
  \Delta^{\mathrm{ans}}_t
  =
  \log \widetilde q_t^a(y_t)
  -
  \log \widetilde \pi_t(y_t),
  \label{eq:answer_only_delta}
\end{equation}
whereas the feedback-conditioned log-ratio is
\begin{equation}
  \Delta^{\mathrm{fb}}_t
  =
  \log \widetilde q_t^{a,f}(y_t)
  -
  \log \widetilde \pi_t(y_t).
  \label{eq:feedback_delta}
\end{equation}
Their difference isolates the contribution of judge feedback:
\begin{equation}
  \delta^{\mathrm{judge}}_t
  =
  \Delta^{\mathrm{fb}}_t
  -
  \Delta^{\mathrm{ans}}_t
  =
  \log \widetilde q_t^{a,f}(y_t)
  -
  \log \widetilde q_t^a(y_t).
  \label{eq:judge_information_layer}
\end{equation}
Equation~\ref{eq:judge_information_layer} makes explicit that judge feedback acts as an additional information layer on top of answer-only supervision.
It does not introduce a separate teacher-generation target; instead, it changes how the teacher scores the same original student completion.

\begin{figure}[htbp]
  \centering
  \includegraphics[width=0.98\linewidth]{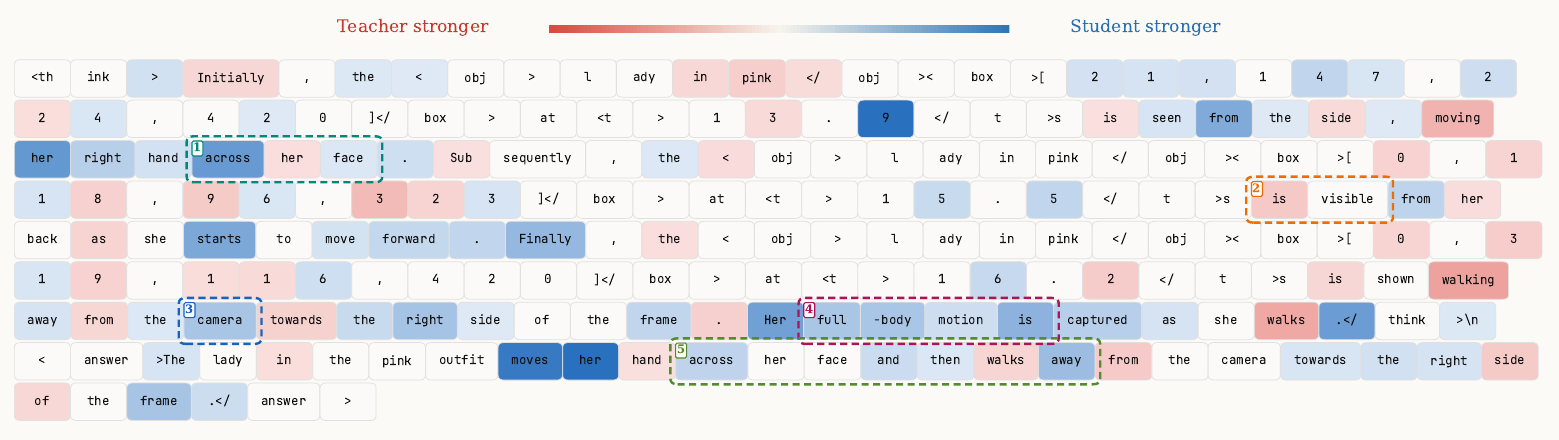}
  \vspace{-0.2em}

  {\small \textbf{(a)} Answer-only teacher replay}

  \vspace{0.8em}
  \includegraphics[width=0.98\linewidth]{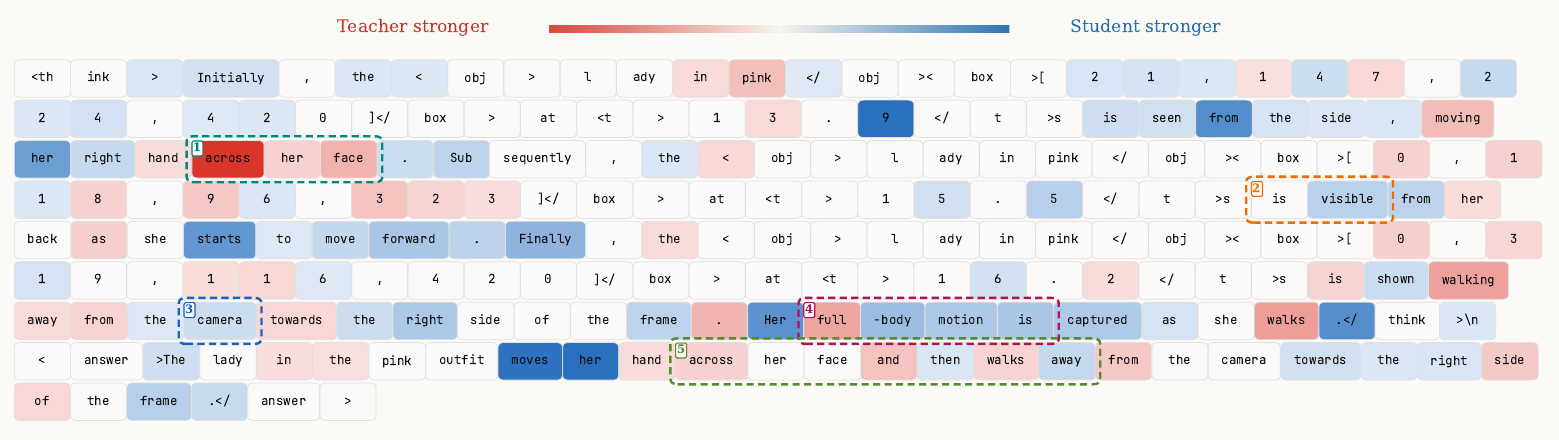}
  \vspace{-0.2em}

  {\small \textbf{(b)} Feedback-conditioned teacher replay}

  \caption{\small \emph{Answer-only versus feedback-conditioned token credit.}
  \textbf{(a)} and \textbf{(b)} visualize the same fixed student rollout; only the teacher context changes.
  Warmer and cooler token backgrounds indicate positive or negative teacher-student token evidence used to modulate policy-gradient magnitude.
  In the answer-only replay, the teacher is conditioned on verified answer-side information, so the token-credit pattern mainly reflects whether the generated trajectory is compatible with the final answer.
  In the feedback-conditioned replay, the teacher additionally receives the judge diagnosis of the rollout, which specifies process-level issues in the reasoning trajectory.
  The important comparison is therefore not whether every token changes, but whether feedback changes token evidence at reasoning, grounding, and answer-relevant positions.
  The numbered outlines mark representative process-relevant positions affected by feedback: \textbf{1} marks the early evidence \texttt{across her face}, which changes from student-favored to teacher-favored after feedback; \textbf{2} marks the intermediate visibility claim \texttt{is visible}; \textbf{3} marks later camera evidence; \textbf{4} marks \texttt{full-body motion is}; and \textbf{5} marks the answer phrase \texttt{across her face and then walks away}.
  The most visible changes occur around the action evidence in \textbf{1}, the motion evidence in \textbf{4}, and the answer-level action phrase in \textbf{5}.
  Most other tokens remain close to the answer-only pattern, indicating that feedback selectively changes process-relevant credit rather than uniformly shifting the entire trajectory.
  This illustrates the intended mechanism of VISD: the rollout-level advantage still determines the update direction, whereas judge feedback adds an information layer that redistributes token-level update magnitudes toward process-relevant parts of the same trajectory.}
  \label{fig:appendix_answer_feedback_heatmap}
\end{figure}

VISD applies this signal through a positive token-wise reweighting factor.
As shown in Appendix~\ref{app:topk_local_support_ratio}, the judge-conditioned teacher can change the relative magnitude assigned to different tokens, but it cannot flip the rollout-level reinforcement direction.
This property is important for our interpretation: feedback contributes diagnostic information for \emph{which process-relevant tokens} should receive different update magnitudes, while the reward advantage still decides whether the trajectory is promoted or suppressed.
Figure~\ref{fig:appendix_answer_feedback_heatmap} should therefore be read as a controlled comparison: the student rollout, reward, and displayed token sequence are fixed, and only the teacher-side context changes from answer-only information to feedback-conditioned diagnostic context.
The visual differences between \textbf{(a)} and \textbf{(b)} instantiate $\delta^{\mathrm{judge}}_t$ in Equation~\ref{eq:judge_information_layer}: they show where the added feedback context changes the teacher-side token evidence used for magnitude reweighting.

\clearpage
\section{Qualitative Case Studies}
\label{app:qualitative_case_studies}
\begin{figure}[h]
  \centering
  \includegraphics[width=\linewidth]{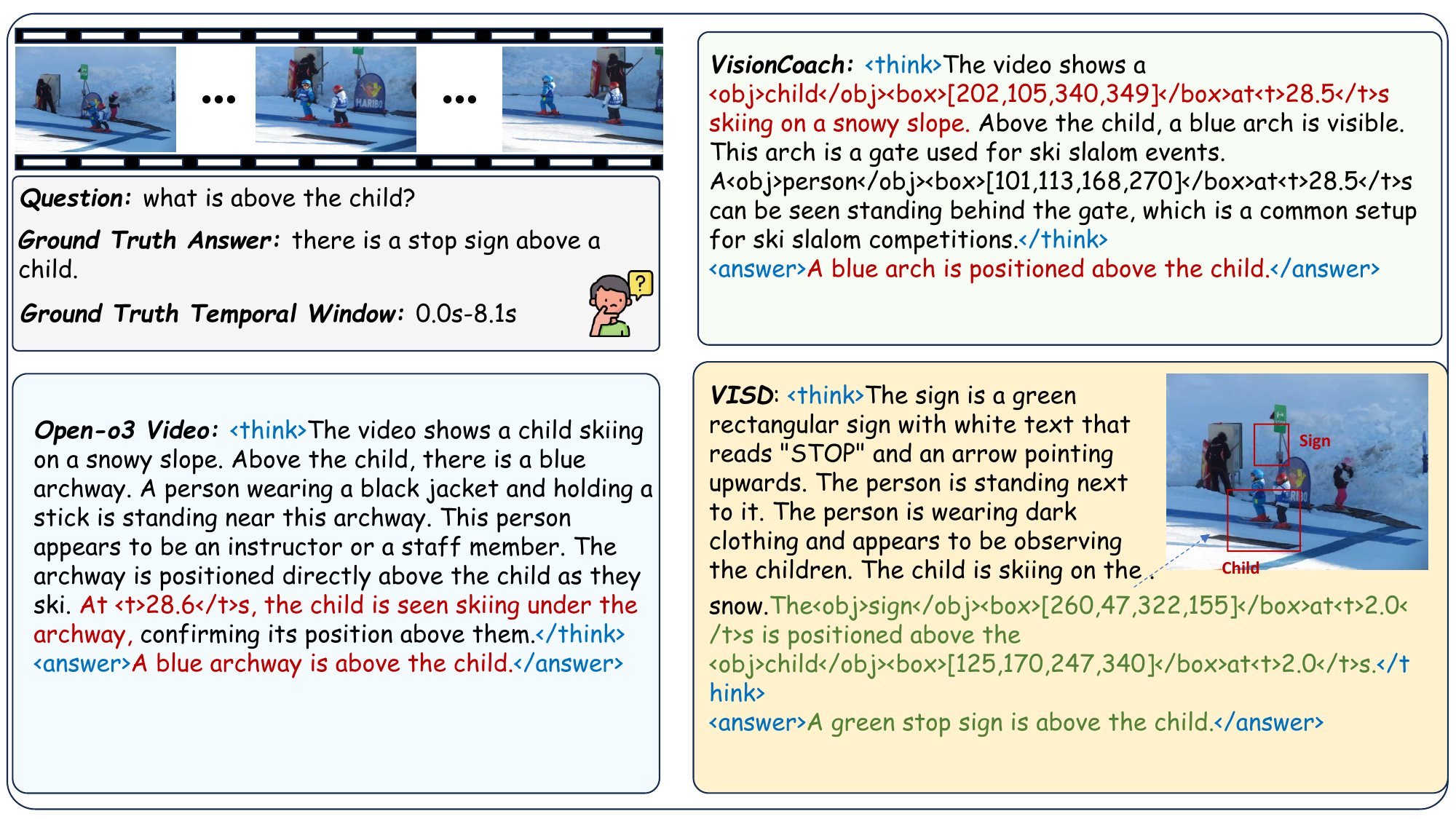}
  \caption{\emph{Visualization}. For spatial relation reasoning, VISD accurately localizes the queried child and identifies the object positioned above him, providing precise visual evidence while avoiding confusion with nearby objects. In contrast, related video reasoning models either give incorrect answers or rely on incomplete spatial grounding.}
  \label{vstar_stop_sign}
\end{figure}

\begin{figure}[h]
  \centering
  \includegraphics[width=\linewidth]{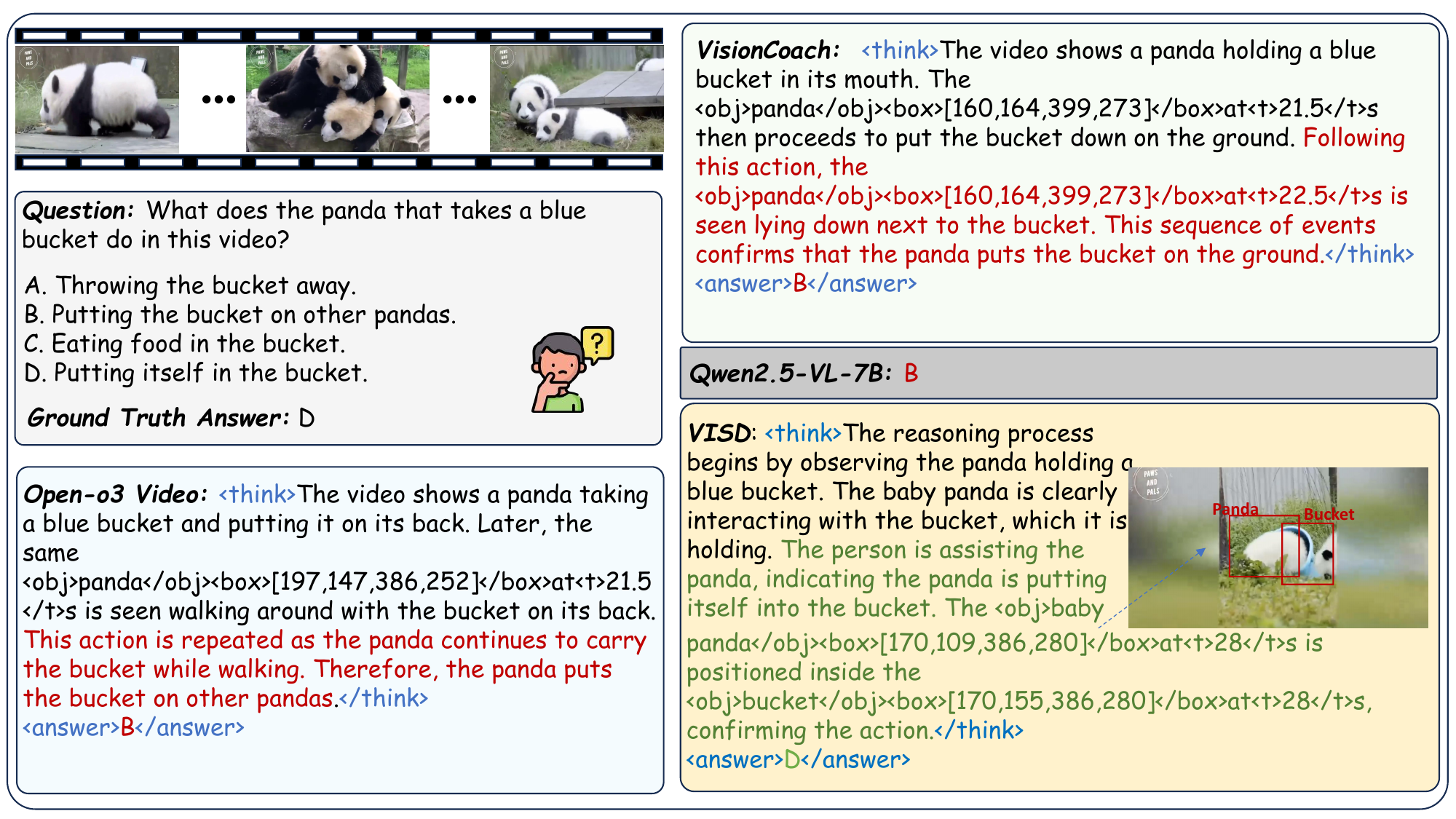}
  \caption{\emph{Visualization}. For temporal action reasoning, VISD grounds the panda and bucket across relevant frames and correctly infers that the panda is putting itself in the bucket. Competing models miss this action transition and produce incorrect answers.}
  \label{mme_panda}
\end{figure}

\begin{figure}[h]
  \centering
  \includegraphics[width=\linewidth]{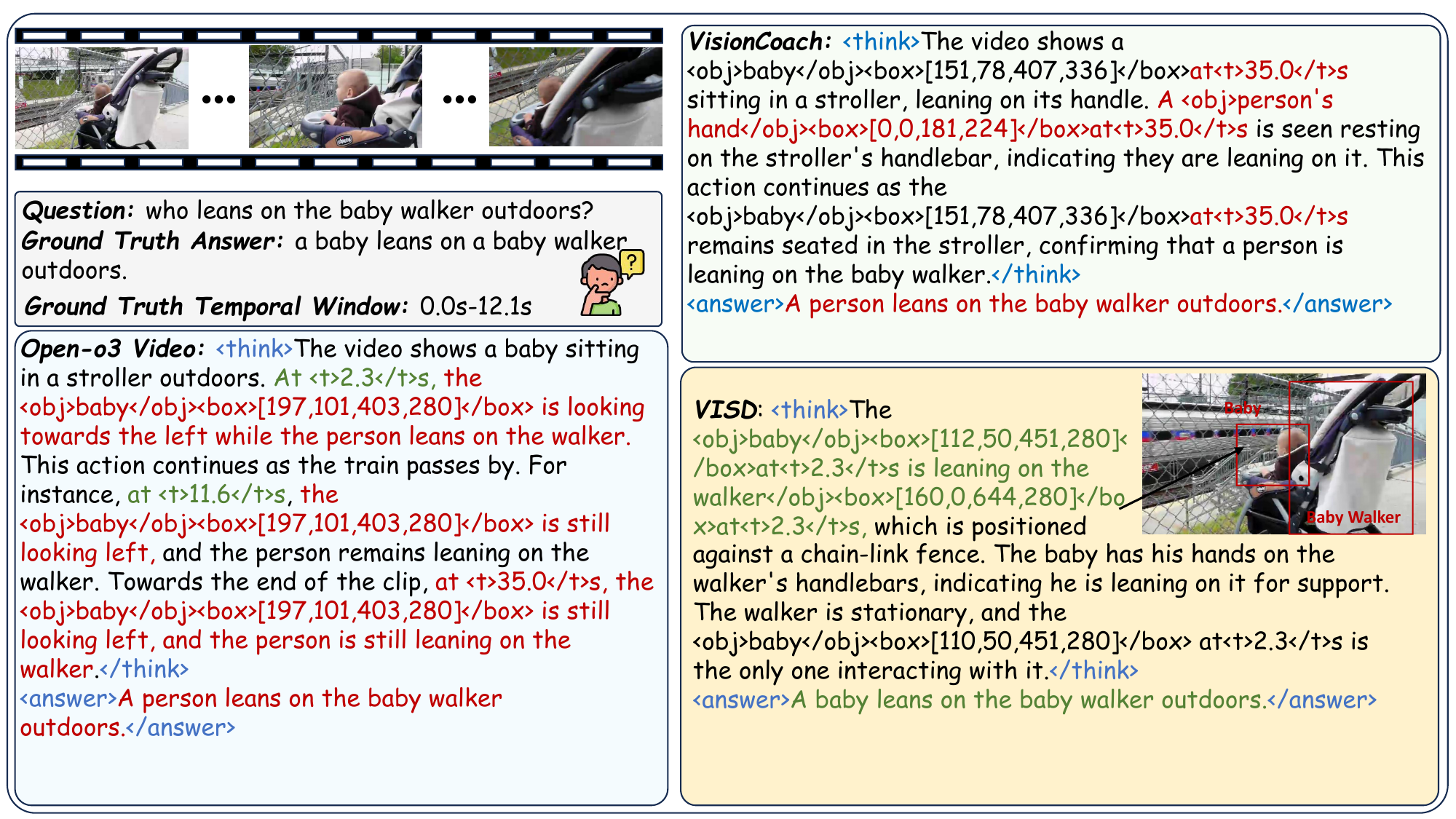}
  \caption{\emph{Visualization}. For outdoor spatial interaction reasoning, VISD accurately localizes both the baby and the baby walker, identifies the baby as the one leaning on the walker, and provides precise visual evidence grounded in the relevant temporal window. Other models instead confuse the person with the baby and produce an incorrect answer.}
  \label{vstar_baby}
\end{figure}

\begin{figure}[h]
  \centering
  \includegraphics[width=\linewidth]{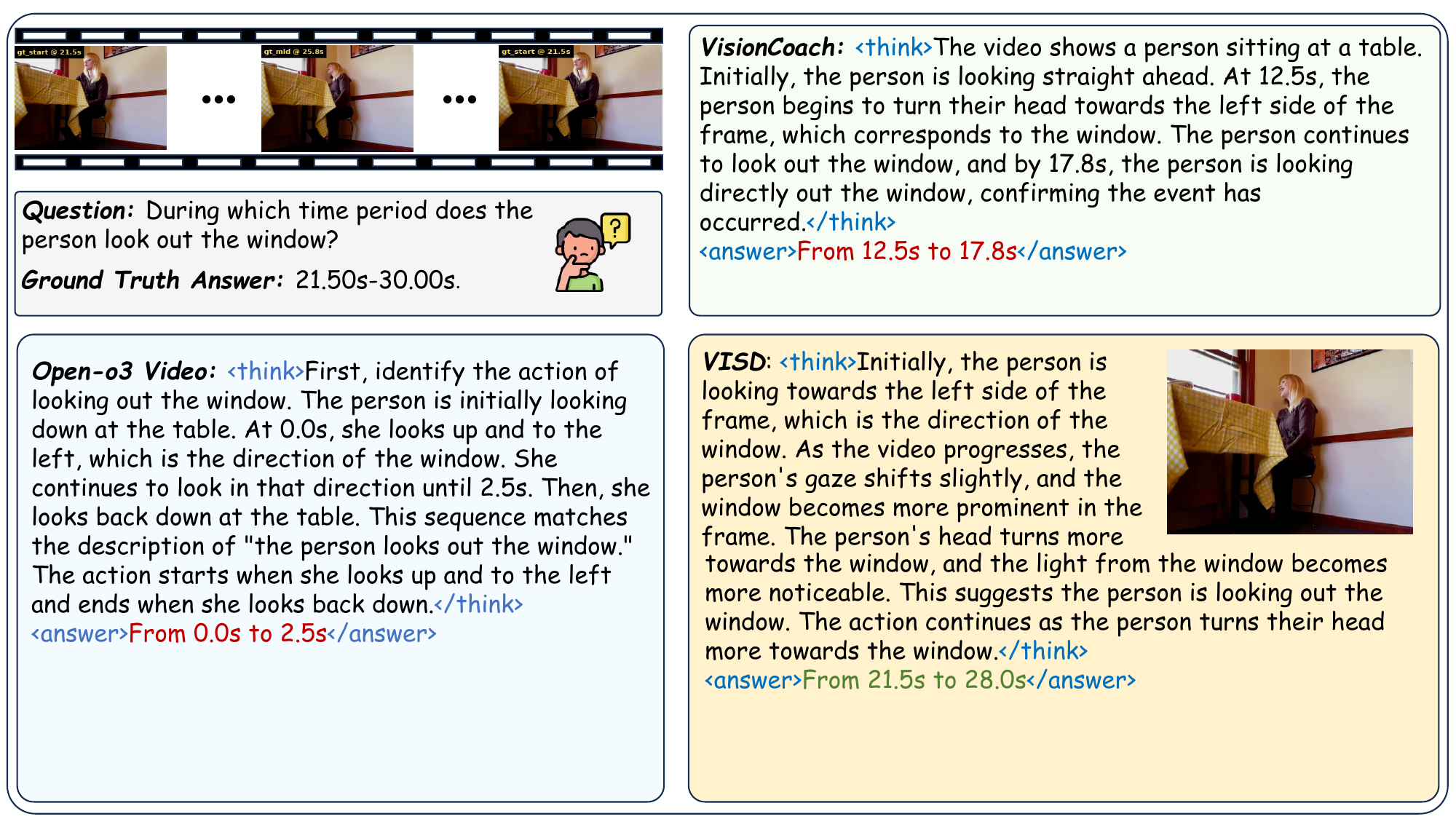}
  \caption{\emph{Visualization}. For temporal localization reasoning, VISD accurately grounds the person's gaze direction across frames and identifies the correct time window when the person looks out of the window. Other models focus on earlier or incomplete head movements and produce incorrect intervals, whereas VISD captures the sustained window-looking action and matches the ground-truth period.}
  \label{case_chara_window}
\end{figure}

\begin{figure}[h]
  \centering
  \includegraphics[width=\linewidth]{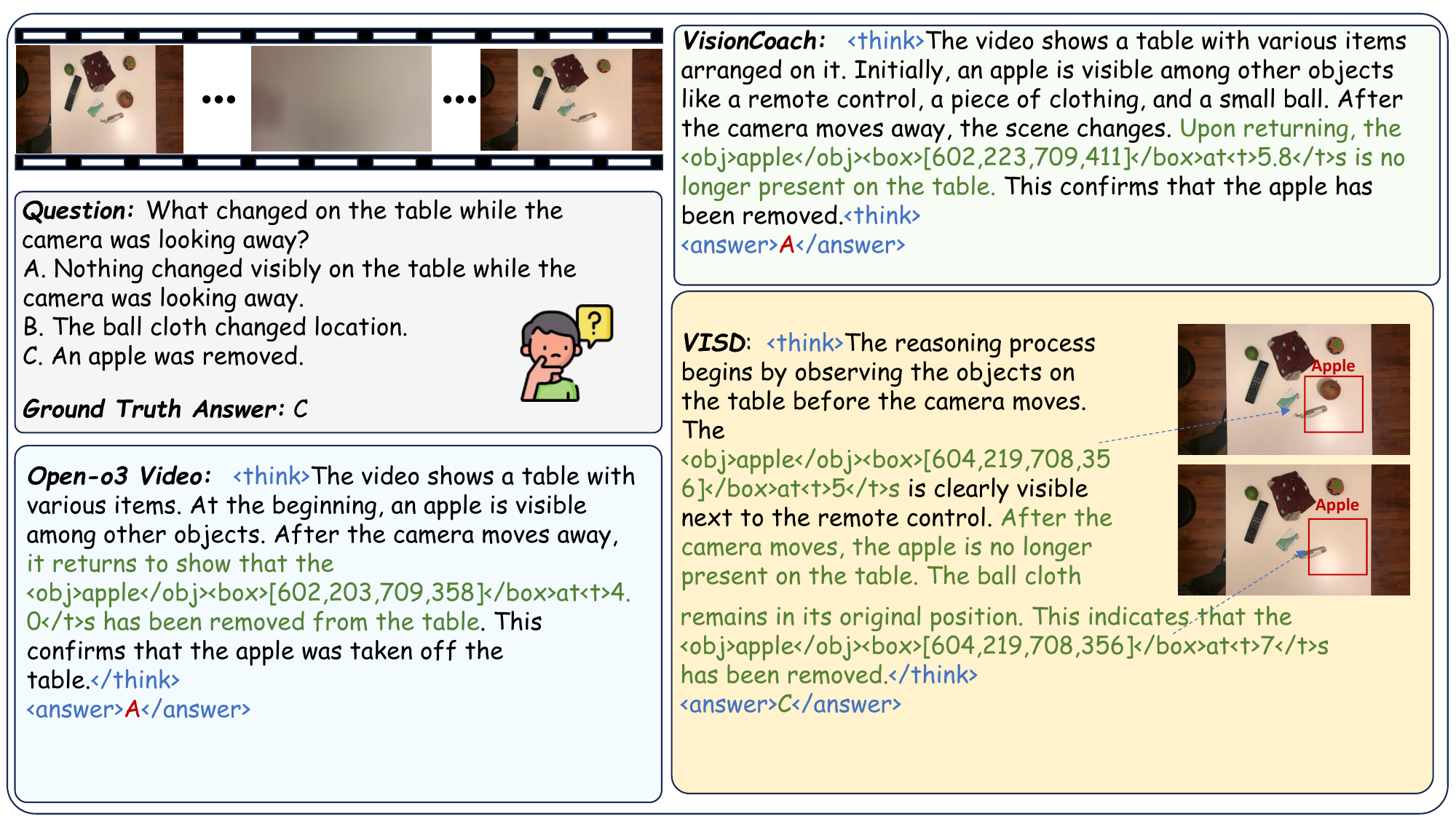}
  \caption{\emph{Visualization}. In object disappearance reasoning, VISD explicitly grounds the apple before and after the camera movement, correctly identifying that the apple has been removed from the table. Although related models partially recognize the same visual change in their reasoning process, they still produce an inconsistent final answer, highlighting VISD's stronger alignment between visual grounding, reasoning, and response generation.}
  \label{case_percep_apple}
\end{figure}

\begin{figure}[h]
  \centering
  \includegraphics[width=\linewidth]{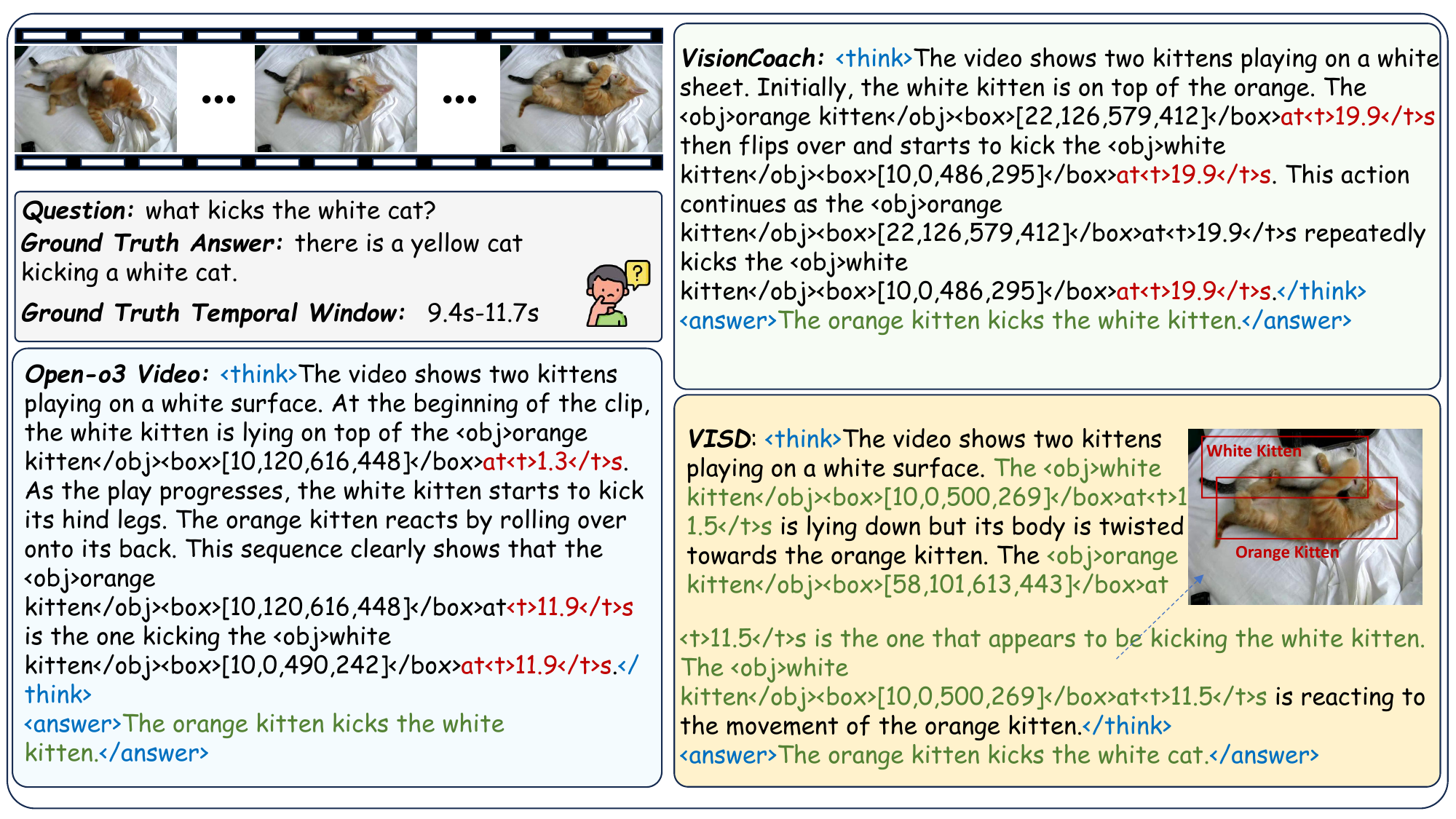}
  \caption{\emph{Visualization}. VISD not only answers correctly but also localizes the action within the correct temporal window, identifying the moment when the orange kitten kicks the white kitten. Although related video reasoning models produce the correct final answer, they rely on inaccurate temporal evidence.}
  \label{case_vstar_cat}
\end{figure}

\begin{figure}[h]
  \centering
  \includegraphics[width=\linewidth]{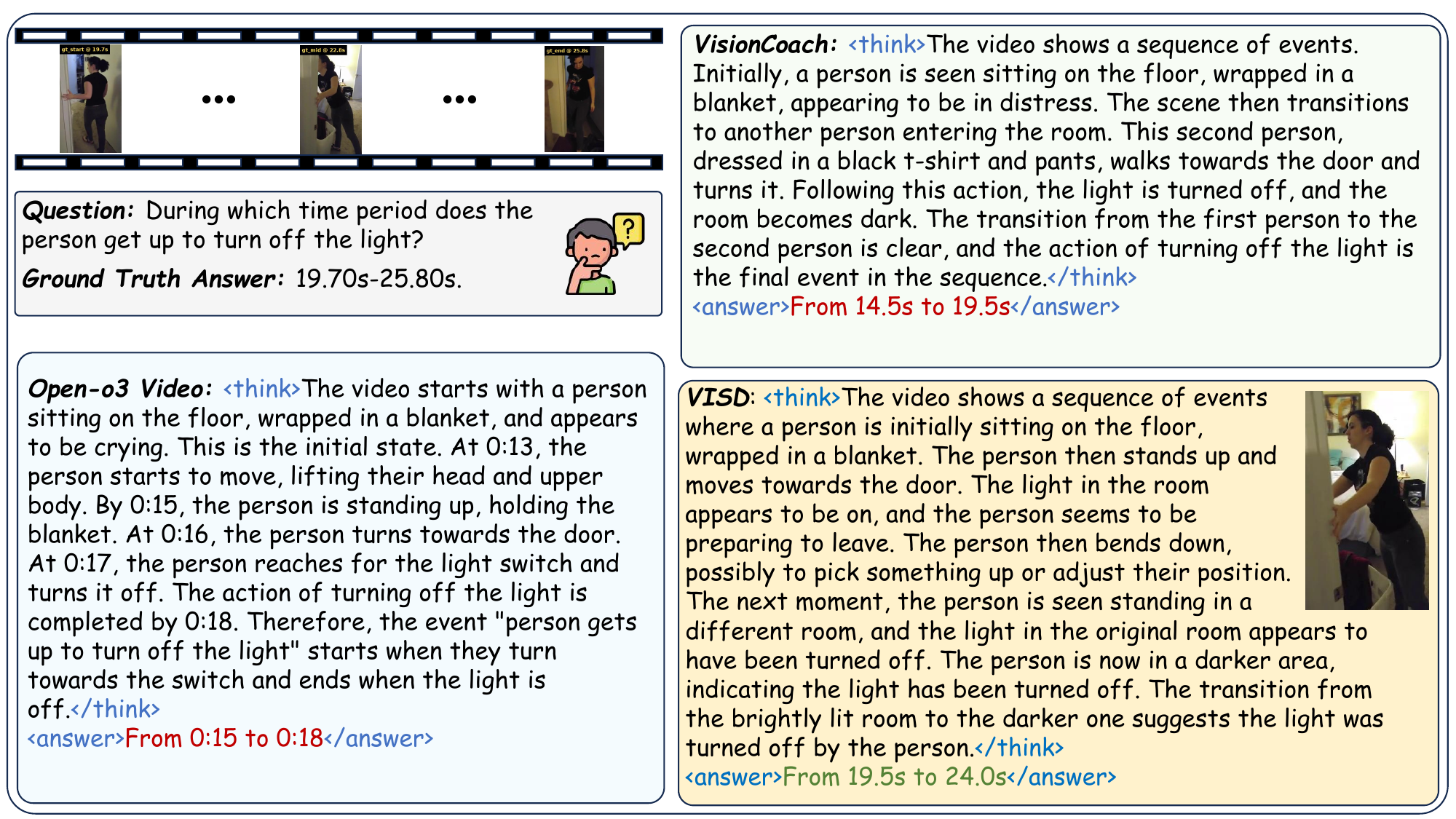}
  \caption{\emph{Visualization}. For temporal action localization reasoning, VISD accurately grounds the key transition from the lit room to the dark room and identifies the correct time interval when the person gets up to turn off the light. In contrast, related video reasoning models focus on earlier incomplete movements, such as sitting up or approaching the switch, resulting in incorrect temporal predictions.}
  \label{case_chara_light}
\end{figure}

\begin{figure}[h]
  \centering
  \includegraphics[width=\linewidth]{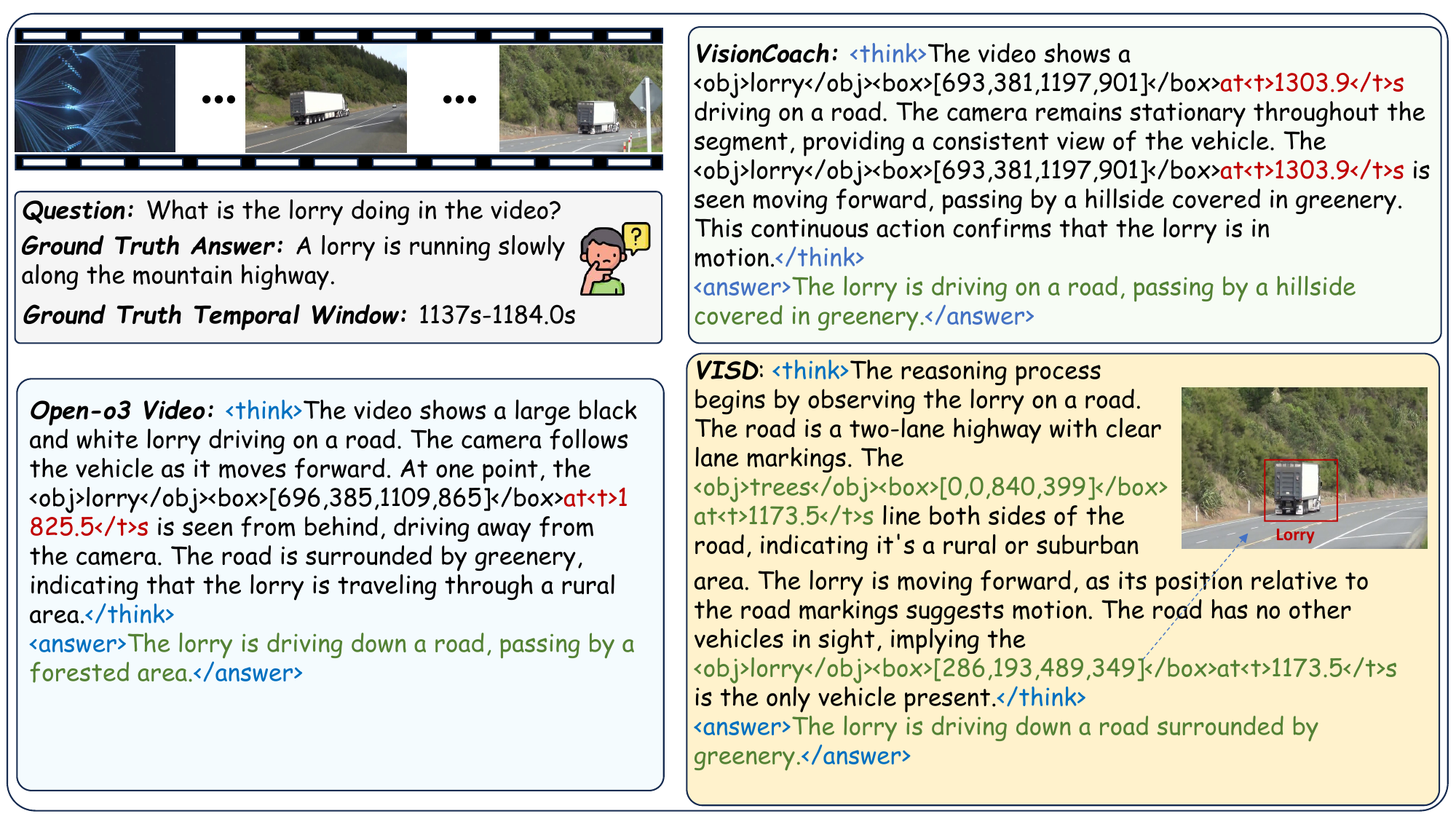}
  \caption{\emph{Visualization}. VISD correctly answers the question and localizes the lorry within the ground-truth temporal window. Although other models also produce the correct answer, they ground their reasoning in an incorrect time interval, showing less precise temporal localization.}
  \label{case_vstar_lorry}
\end{figure}

\begin{figure}[h]
  \centering
  \includegraphics[width=\linewidth]{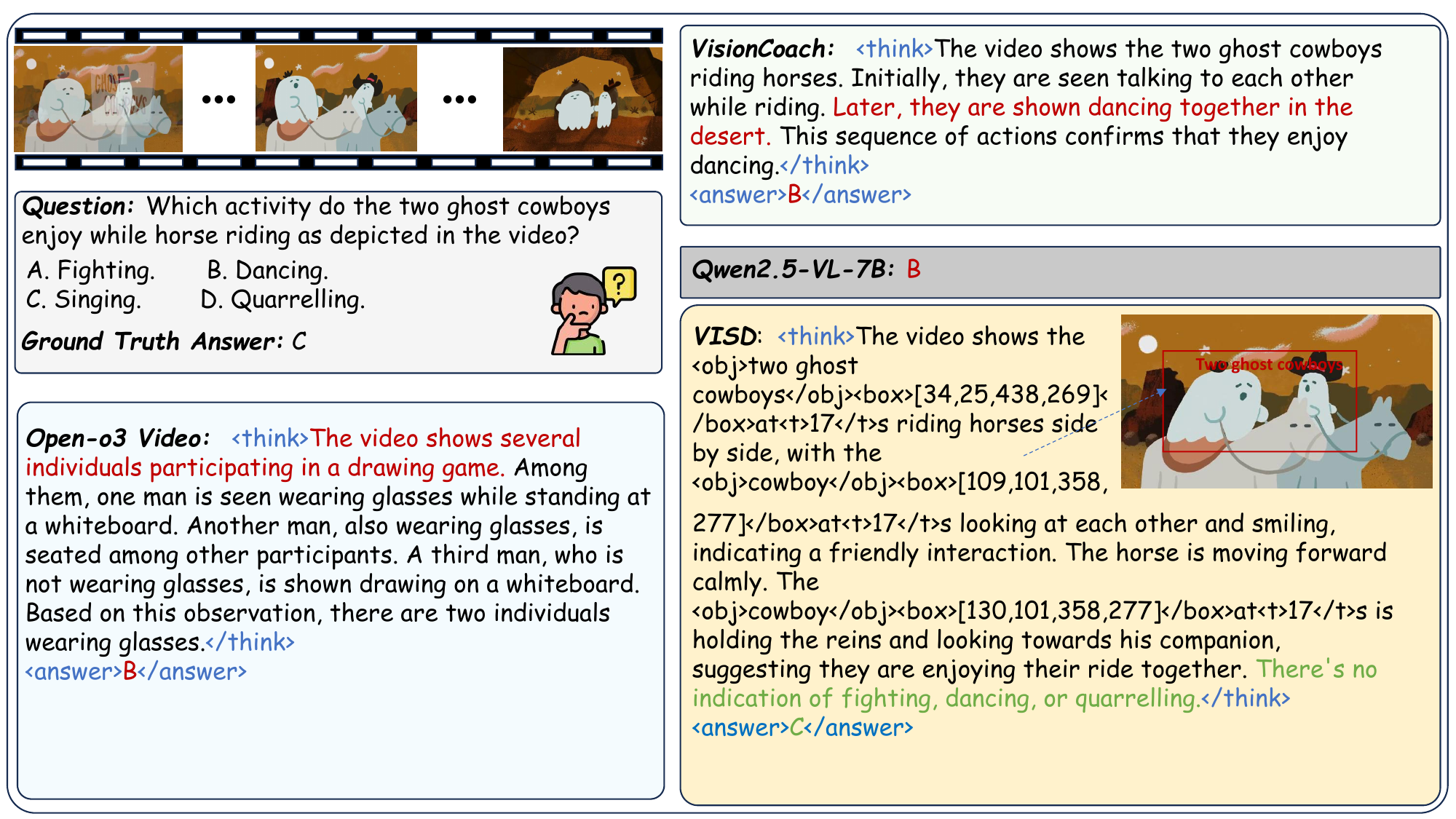}
  \caption{\emph{Visualization}. In fine-grained activity reasoning, VISD accurately grounds the two ghost cowboys during horse riding and identifies their smiling, friendly interaction, correctly inferring that they enjoy singing rather than dancing, fighting, or quarrelling. In contrast, competing models are distracted by later or irrelevant visual cues and produce incorrect answers.}
  \label{case_mme_ghost}
\end{figure}

To further illustrate the effectiveness of VISD, we provide qualitative case studies covering diverse video understanding scenarios.
As shown in Figures~\ref{vstar_stop_sign}--\ref{case_mme_ghost}, VISD grounds its reasoning in precise visual evidence, localizing relevant objects, actors, actions, and temporal intervals before producing the final answer.
Compared with existing video reasoning models, which often rely on incomplete observations or incorrect temporal and spatial evidence, VISD demonstrates stronger alignment between visual grounding, intermediate reasoning, and response generation.
These examples show that VISD not only answers correctly, but also provides more faithful and interpretable reasoning processes across both spatial and temporal video understanding tasks.

\newpage

\clearpage
\section{Judge Prompt Template}
\label{app:judge_prompt}

We provide the core prompt template used by the video-aware judge to generate structured privileged feedback.
The template contains a fixed system instruction and the user prompt assembled from the task type, reference answer, optional grounding evidence, and the student rollout.
Placeholders such as \texttt{\{question\}} and \texttt{\{student\_output\}} denote fields filled at training time.

\refstepcounter{subsection}
\addcontentsline{toc}{subsection}{\protect\numberline{\thesubsection}System Prompt}

\begin{promptbox}{System Prompt: Video-Aware Judge}
## Role
You are a strict video and image QA judge.

## Mission
You are judging a student model's response. Your feedback will be given to a teacher model, which will use it to better guide the student model. Judge whether the student's final answer is correct, whether the student's reasoning broadly supports that final answer, and what the single most likely high-level source of error is when the response is not fully correct.

## Input Boundaries
- Use only the information provided in the prompt.
- You do not see the original video frames directly.
- Do not pretend to verify frame-by-frame visual alignment that is not supported by the provided text or hidden grounding evidence.
- For structured answers, compare the provided fields against the reference answer and grounding evidence when available.
- Treat semantically equivalent natural-language answers as correct even if the wording differs.

## Output Format
- Return strict JSON only with key: feedback.
- Do not return markdown, prose outside JSON, or extra keys.

## Feedback Requirements
- The feedback should be concise, factual, correction-oriented, and usually two to four sentences.
- Mention only the issues that actually appear; do not give an exhaustive recap.
- If the response is fully correct and no clear issue is observed, provide a brief positive feedback sentence stating that the final answer is correct and the reasoning broadly supports it.

## Evaluation Focus
1. Answer diagnosis.
Say whether the student final answer is fully correct, partly correct, or wrong. Briefly name the problematic part. For natural-language answers, focus on semantic meaning rather than exact wording. For structured answers, say which part is wrong or missing.
2. Reasoning-versus-answer consistency.
Judge whether the student's reasoning broadly supports the final answer. If the reasoning points to one event, object, text clue, time range, or spatial reference but the final answer states another, say so. If the reasoning is too weak, too broad, or too incomplete to justify the final answer, say that clearly.
3. High-level error cause.
When the response is not fully correct, pick the single most likely high-level cause and mention it briefly. Prefer one of these categories: the reasoning focused on the wrong event, time span, or object; the reasoning was too broad or lacked enough evidence to support such a specific answer; the reasoning was mostly on the right track but the final answer overreached, drifted, or stated the wrong thing; or the output was incomplete and never produced a real final answer. Do not invent fine-grained step-by-step visual mistakes when they are not well supported by the text.

## Task
Compare the verified answer against the student's final answer and reasoning text, then return the feedback JSON.

If keyframe object evidence is provided:
Hidden grounding evidence may be provided in the user prompt.

If keyframe object evidence is not provided:
No hidden grounding evidence is required for this sample.

If the student output does not contain a final <answer> or </answer> tag:
The student output does not contain a final <answer> or </answer> tag. You must still analyze the student's reasoning text and provide corrective feedback. In the feedback, explicitly point out that the final answer was not produced and that the response may have been truncated because the thinking section was too long. Treat the missing final answer as an incomplete response.
\end{promptbox}

\newpage
\refstepcounter{subsection}
\addcontentsline{toc}{subsection}{\protect\numberline{\thesubsection}User Prompt Assembly}

\begin{promptbox}{User Prompt: Runtime Assembly}
Please judge the following sample.

## Task And Reference Context
- Task type: {task}
- Question: {question}
- Standard answer: {standard_answer}
- Keyframe object evidence: {keyframe_object_evidence_json_or_null}
- Keyframe object evidence is hidden grounding reference. Use it only to judge whether the student's reasoning points to the right time, object, and spatial region.

## Student Submission
- Student final answer: {model_answer_or_null_no_answer_tag_found}
- Student full output: {student_output}

## Required Output
{
  "feedback": "{concise diagnostic feedback}"
}
\end{promptbox}

\section{Broader Impacts}
\label{app:broader_impacts}

VISD aims to improve video understanding systems in tasks that require more reliable reasoning over temporal and spatial evidence.
Such capability can be beneficial for applications including video retrieval, assistive video understanding, educational tools, and human-in-the-loop video analysis, where better grounding may improve interpretability and reduce obvious reasoning errors.

At the same time, stronger video reasoning models may also introduce risks.
They could be used in surveillance-related settings, content analysis pipelines, or other scenarios where incorrect or overconfident predictions may affect downstream decisions.
Like other large multimodal models, they may also inherit biases from pretraining data or produce misleading grounded explanations when the underlying prediction is wrong.
In this work, we evaluate VISD only on public research benchmarks and release code for research purposes, and we view responsible deployment, dataset governance, and more robust evaluation under open-world settings as important future directions.

\clearpage

\end{document}